\documentclass{article}
\usepackage{float}
\usepackage{PRIMEarxiv}

\usepackage[utf8]{inputenc} 
\usepackage[T1]{fontenc}    
\usepackage{hyperref}       
\usepackage{url}            
\usepackage{booktabs}       
\usepackage{amsfonts}       
\usepackage{nicefrac}       
\usepackage{microtype}      
\usepackage{lipsum}
\usepackage{graphicx}
\graphicspath{{media/}}     
  
\title{NexViTAD: Few-shot Unsupervised Cross-Domain Defect Detection via Vision Foundation Models and Multi-Task Learning
\thanks{
The source code for NexViTAD is publicly accessible on: https://github.com/mutianwei521/NexViTAD.}} 

\author{
  Tianwei Mu$^*$\\
    \textit{Guangzhou Institute of Industrial Intelligence}\\
  Guangzhou\\
  1189233@mail.dhu.edu.cn\\
   \And
  Feiyu Duan, Bo Zhou\\
  \textit{School of Architecture \& Civil Engineering}\\
  Shenyang University of Technology\\
  Shenyang\\ \\
     \And
  Dan Xue\\
  School of Information Science and Engineering\\
  Shenyang University of Technology\\
  Shenyang\\
     \And
  Manhong Huang\\
  College of Environmental Science and Engineering\\
  \textit{Donghua University}\\
  Shanghai\\
}

\begin{document}
\maketitle

\begin{abstract}
This paper presents a novel few-shot cross-domain anomaly detection framework, Nexus Vision Transformer for Anomaly Detection (NexViTAD), based on vision foundation models, which effectively addresses domain-shift challenges in industrial anomaly detection through innovative shared subspace projection mechanisms and multi-task learning (MTL) module. The main innovations include: (1) a hierarchical adapter module that adaptively fuses complementary features from Hiera and DINO-v2 pre-trained models, constructing more robust feature representations; (2) a shared subspace projection strategy that enables effective cross-domain knowledge transfer through bottleneck dimension constraints and skip connection mechanisms; (3) a MTL Decoder architecture supports simultaneous processing of multiple source domains, significantly enhancing model generalization capabilities; (4) an anomaly score inference method based on Sinkhorn-K-means clustering, combined with Gaussian filtering and adaptive threshold processing for precise pixel level. Evaluated on the MVTec AD dataset, NexViTAD delivers state-of-the-art performance with an AUC of 97.5\%, an AP of 70.4\%, and a PRO of 95.2\% in the target domains, surpassing other recent models, marking a transformative advance in cross-domain defect detection.
\end{abstract}


\section{Introduction}
    Automated defect detection plays an essential role in ensuring the quality and safety of industrial manufacturing. Traditional supervised deep learning methods have demonstrated strong performance in detecting surface anomalies of objects. However, their success is critically based on large-scale, manually annotated datasets, an unrealistic assumption in many real-world settings where defect samples are extremely rare and expensive to acquire \cite{Tsai_2018}. Furthermore, these models often suffer significant performance degradation when deployed in new environments due to domain shifts caused by variations in materials, illumination, sensors, or production workflows \cite{batzner2024efficientad,Str_ter_2024}.
    
    To mitigate these issues, unsupervised anomaly detection (UAD) has received increasing attention that reduces the annotation burden, but they still struggle to transfer knowledge across domains and often lack the semantic richness required for fine-grained localization of complex defects.
    
    Cross-domain learning has emerged as a promising approach \cite{Bergmann_2019,Lai_2022,he2024anomalycontrol}. In this paradigm, models trained on a labeled, defect-rich source domain are adapted to an unlabeled or weakly labeled target domain that differs in sensor types, lighting conditions, product materials, or image acquisition setups. By aligning features across domains, cross-domain approaches help models learn domain-invariant representations and semantic correlations that transfer better across contexts.
    
    In fact, cross-domain strategies have demonstrated improved anomaly detection accuracy by reducing the false alarm rate and boosting robustness in new operational environments. However, they also come with notable limitations \cite{Fatima_2024, Hsu_2025,Li_2024a,Liu_2025,He_2025}: (1) They struggle to adapt to various types of defect across different industrial objects, often failing to generalize due to domain-specific feature extraction; (2) Their reliance on rigid, source-domain-only supervision hampers adaptability, resulting in degraded performance on unlabeled target domains; (3) They often lack the precision required for pixel-level defect localization, particularly under domain shifts, limiting their effectiveness in complex industrial scenarios.

\begin{figure}[h]
        \centering
        \includegraphics[width=12cm]{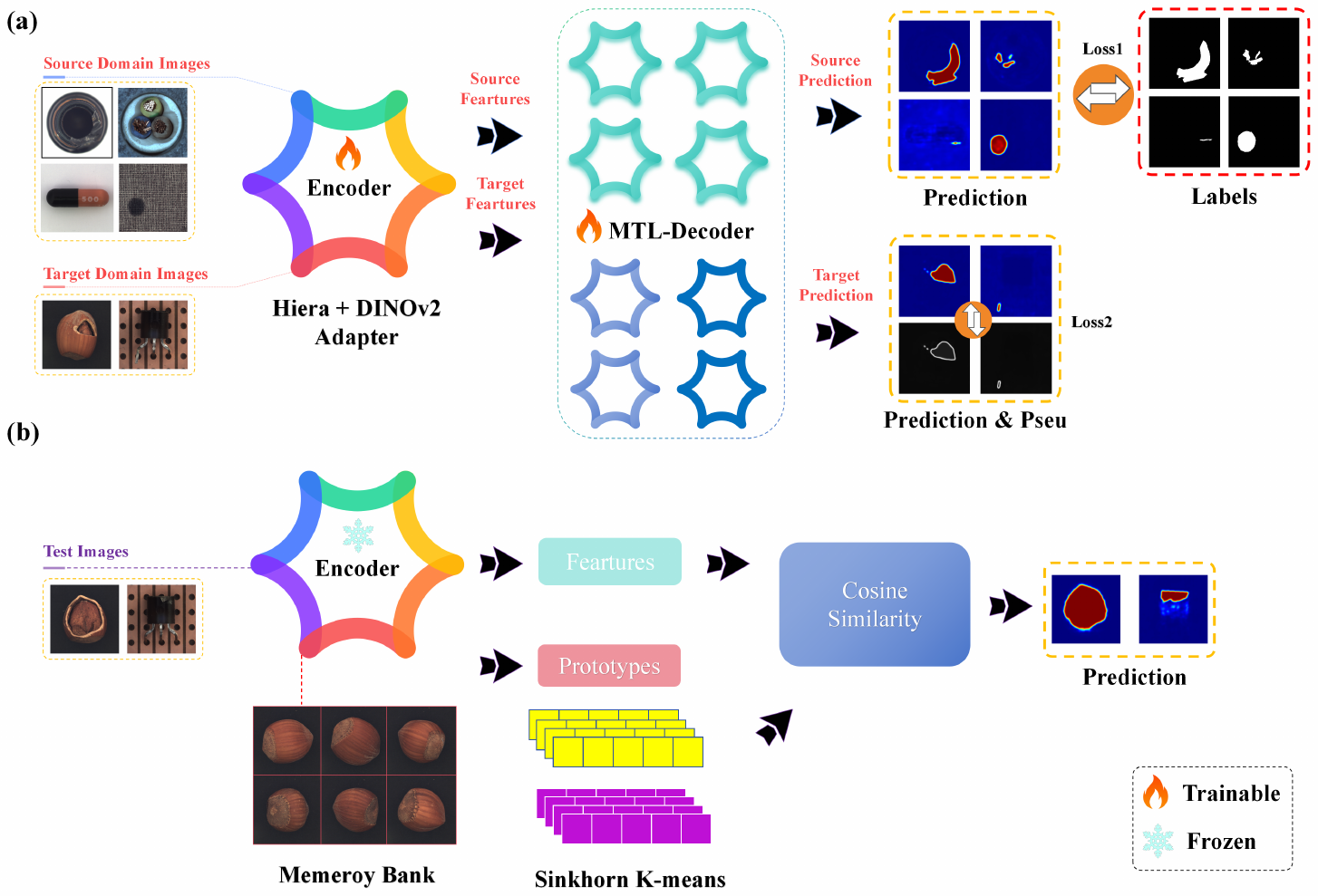}
        \caption{Overview of the NexViTAD framework: (a) showing the Encoder, multi-task Decoder, and (b) inference pipeline with memory bank clustering}
        \label{fig:1}
    \end{figure}

    To overcome these challenges, vision foundation models such as DINOv2 \cite{oquab2023dinov2} and Hiera \cite{ryali2023hiera} provide powerful pretrained representations, yet their integration into cross-domain unsupervised defect detection with fine-grained localization remains underexplored. Inspired by recent work in medical imaging \cite{zhang2025adapting}, we propose NexViTAD, a novel framework designed to adapt to various types of defects in industrial objects, reduce dependency on source domain supervision, and achieve precise location of pixel levels between domains. NexViTAD leverages the Hiera Encoder for adaptive multi-scale feature extraction across varied object types, integrates DINOv2 features for enhanced semantic robustness, and employs a multi-task learning (MTL) Decoder with task-specific heads for source domain classification and confidence-based pseudo-labeling to enable effective adaptation to unlabeled target domains. During inference, NexViTAD removes the Decoder and uses a memory bank of normal samples with Sinkhorn K-means clustering to deliver accurate pixel-level anomaly localization. \textbf{Figure \ref{fig:1} }illustrates the NexViTAD framework.
    \raggedbottom
    
    Our main contributions are as follows: (1) NexViTAD, a novel framework that integrates hierarchical vision transformers and self-supervised DINOv2 features for cross-domain unsupervised defect detection; (2) A pseudo-labeling strategy with confidence thresholding and perturbation consistency for robust target domain learning; (3) An efficient inference pipeline using memory banks and Sinkhorn K-means for unsupervised anomaly detection; (4) State-of-the-art performance on the MVTec AD dataset, with extensive ablation studies validating each component.

\section{Related Work}
\label{sec:headings}
    Industrial defect detection has made substantial strides in recent years, driven by advances in deep learning. Four major methodological categories with references: supervised anomaly detection, unsupervised anomaly detection, cross-domain few-shot learning, and self-supervised/multi-task learning.
    \subsection{Supervised Anomaly Detection}
        Traditional defect detection methods rely on supervised convolutional neural networks (CNNs) such as UNet \cite{Ronneberger_2015} or ResNet \cite{He_2016}, which require large labeled datasets and struggle with domain shifts \cite{Str_ter_2024}. The UNet’s Encoder–Decoder architecture excels in pixel‑level localization, while ResNet's deep residual connections enhance representation learning. However, their reliance on large labeled datasets, especially scarce defect annotations, limits scalability. Moreover, these models degrade significantly under minor domain shifts (e.g., changes in lighting, sensor types, materials), indicating poor generalization in complex industrial settings \cite{yang2025survey}. These limitations restrict their applicability in real-world industrial settings, where defect samples are rare.
    \subsection{Unsupervised Anomaly Detection}
        To reduce dependence on labeled data, unsupervised anomaly detection (UAD) techniques learn from non-defective (normal) samples only, identifying deviations at inference time. Recent approaches, such as PatchCore \cite{Roth_2022}, have constructed a memory bank of normal feature patches. Anomalies are detected by measuring distances between test patches and their nearest neighbors, which achieves strong performance on datasets such as MVTec AD. FastFlow \cite{yu2021fastflow} leverage unsupervised anomaly detection by modeling normal sample distributions. EfficientAD \cite{Thakran_2023} employs a student-teacher formulation with objective consistency of features, achieving real-time speeds. SimpleNet and SuperSimpleNet \cite{Liu_2023,Rolih_2024} propose compact architectures that combine embedding and reconstruction tasks in a streamlined pipeline, reducing complexity without sacrificing precision. NSA‑N and NSA‑M \cite{Li_2015} generate natural and man-made anomalies, respectively, through image perturbations. They enhance data diversity, but also introduce artifacts that may confuse detectors. CutPaste \cite{Li_2021} synthetically generates anomalies by randomly cutting and pasting patches within images. DRAEM \cite{Zavrtanik_2021} introduces Perlin noise-based pseudo-defects to train denoising networks. It improves visual realism but generates limited shape variety. DDAD \cite{Mousakhan_2025} integrates diffusion models to perform anomaly detection, enabling pixel-level to generate anomaly localization. RealNet \cite{Zhang_2024} and GLASS \cite{Chen_2024} employ adversarial and global-local co-regulated generation frameworks, producing more realistic anomalies controlled by global and local attributes. However, many UAD methods rely on frozen backbones and fail to adapt to new domains, limiting pixel-wise detection under unseen conditions.
    \subsection{Cross-Domain Few-shot Learning}    
        Industrial environments frequently subject inspection systems to domain shifts due to changes in materials, production environments, or devices. To maintain robustness under limited supervision, cross-domain few-shot learning methods like XDNet \cite{Lee_2023} proposes a meta-learning framework tailored for cross-domain visual inspection, formulated as an n-way, k-shot classification problem to handle image-level defect identification across heterogeneous domains. It enhances generalization by integrating squeeze-excitation modules, anti-aliasing filters, and a contrastive learning loss during meta-training, followed by a non-parametric, Sinkhorn K-means classifier \cite{zhang2023detect} with Self-Optimal Transport at inference time. CAI-N and CAI-M \cite{Shi_2025} use cross-domain anomaly injection to transplant real-world defect patterns into target images using Poisson blending and diffusion techniques. Despite effectiveness at the image level, most methods still struggle with fine-grained localization in heterogeneous environments.
    \subsection{Self-Supervised Pretraining and Multi-Task Learning} 
        Self-supervised learning (SSL) and multi-task learning (MTL) have recently been applied to anomaly detection to improve robustness and enable task transfer. DINOv2 \cite{oquab2023dinov2} is a vision transformer-based SSL model that employs self-distillation to learn domain-agnostic visual representations. It has demonstrated strong performance in transfer learning tasks, including segmentation and anomaly detection in defect datasets. Hiera \cite{ryali2023hiera} uses hierarchical attention to model global and local features in a self-supervised framework, providing useful multi-resolution representations for downstream tasks. These SSL models have been successfully integrated into UAD pipelines (e.g. PatchCore + DINOv2), improving detection consistency \cite{Damm_2025}. Models like DINOv2 and Hiera provide robust feature representations through self-supervised pre-training. Medical imaging studies, such as heart recognition \cite{zhang2025adapting}, have adapted these models for segmentation tasks, but their application to the detection of industrial defects is limited. Thus, we need to extend these ideas to cross-domain settings.
        
        MTL frameworks combine segmentation and classification tasks with pseudo-labeling to leverage unlabeled data. This concept has proven effective in other domains such as medical imaging, automatic driving, and energy forecasting \cite{Lai_2022,zhang2025adapting,Li_2024, caruana1997multitask}. Recent work applies MTL to medical imaging\cite{Gong_2021}, but integrating MTL with pseudo-labeling for unsupervised domain adaptation remains underexplored.
        
        Therefore, NexViTAD is proposed to bridge these gaps by integrating: (1) Cross-domain few-shot adaptation; (2) Double adapting backbone representations (DINOv2 and Hiera); (3) unsupervised inference; (4) MTL with pseudo-labeling head to transfer supervision into the target domain; (5) Memory-bank anomaly detection at inference via clustering.
    \section{Methodology}        
    Let $I \in \mathbb{R}^{H \times W \times 3}$ be an input image, with $S \in \{0, 1\}^{H \times W}$ as the binary segmentation map (0 for normal, 1 for defective). NexViTAD aims to train on a labeled source domain $\mathcal{D}_s = \{(I_i, S_i)\}_{i=1}^{N_s}$ with $C_s$ classes (e.g., tile, leather) and an unlabeled target domain $\mathcal{D}_t = \{I_i\}_{i=1}^{N_t}$. Then it performs unsupervised inference on an unlabeled target domain $D_t$.
    \subsection{Training Phase}
\begin{figure}[H]
            \centering
            \includegraphics[width=12cm]{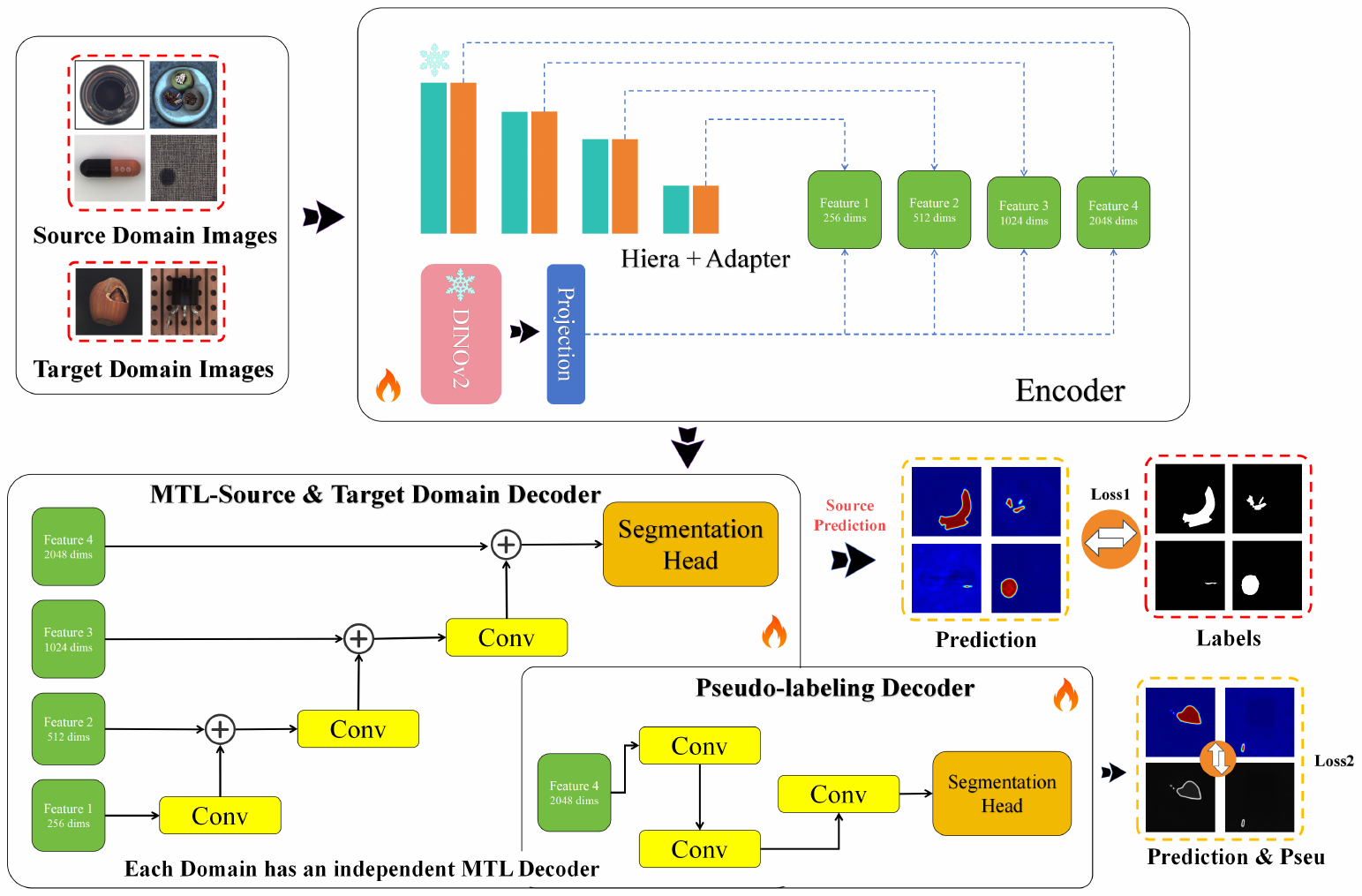}
            \caption{Architecture of the NexViTAD Encoder-Decoder framework, highlighting the Hiera adapter, DINOv2 feature interleaving, and multi-task Decoder with source and target domain heads.}
            \label{fig:2}
        \end{figure}

        NexViTAD adapts the framework from \cite{zhang2025adapting}, with modifications for defect detection, as shown in \textbf{Figure \ref{fig:2}}.
        \raggedbottom
        \subsubsection{Encoder}
            NexViTAD introduces a shared subspace projection Encoder. This mechanism is designed to enable robust cross-domain knowledge transfer by aligning heterogeneous feature representations into a unified dimensional space while preserving both structural detail and semantic richness.
           
            The shared subspace projection module uses the frozen Hiera Encoder \cite{ryali2023hiera} to extract multi-scale features $f_{hiera}(I)=\{F_n^{h} \}_{n=1}^{N}$, where $F_n^h\in \mathbb{R}^{b\ \times \ h_n\times \ w_n\times \ {d_{hiera}}\ }$ represents features at scale $n$, with batch size $b$, spatial dimensions $(h_n,\ w_n)$, and feature dimension $d_{hiera}=[256,\ 512,\ 1024,\ 2048]$ as $N=4$. A lightweight adapter is applied after Hiera's multi-scale attention block in \textbf{Equation \ref{GrindEQ__1_}}, and only adapter parameters are updated during training.
            \begin{equation}\label{GrindEQ__1_}Adapter\left(X\right)=\sigma\left(X_nW_{down}+b_{down}\right)W_{up}+b_{up}+X_n\end{equation}

            where $X_n = F_n^h \in R^{b \times h_n \times w_n \times d_{hiera}}$ is the multi-scale Hiera output; $W_{down} \in \mathbb{R}^{d_{hiera} \times r}$; $b_{down} \in \mathbb{R}^r$; $W_{up} \in \mathbb{R}^{r \times d_{hiera}}$; $b_{up} \in \mathbb{R}^d_{hiera}$; $r = \frac{d_{hiera}}{4}$; $\sigma(\cdot)$ is the GELU activation.

            DINOv2 features $f_{dino}(I) \in \mathbb{R}^{b \times h \times w \times d_{dino}}$ ($d_{dino} = 384$) are extracted without fine-tuning. These are resized to match Hiera’s spatial dimensions ($h_n$, $w_n$) via bilinear interpolation, yielding $F_n^d \in \mathbb{R}^{b \times h_n \times w_n \times d_{dino}}$. A projection layer aligns DINOv2 channels to Hiera:

            \begin{equation}
                \label{GrindEQ__2_}F_n^d=\ F_n^dW_{proj}+b_{proj}
            \end{equation}

            where $W_{proj}\in\mathbb{R}^{d_{dino}\times\ d_{hiera}};\ b_{proj}\in\mathbb{R}_{hiera}^d$

            Then, the features from both sources are interleaved slice-by-slice along the channel dimension. This interleaving ensures that information from both Encoders is spatially and semantically aligned within a common subspace, as expressed in \textbf{Equation \ref{GrindEQ__3_}}. The resulting representation $F_n$ combines the multi-scale locality of Hiera with the semantic abstraction of DINOv2 in a shared latent space, enabling the Decoder to leverage both during downstream tasks. The final multi-scale representation is $F={\{F_1,\ \ldots,\ F_N\}}$.
            
            \begin{equation} \label{GrindEQ__3_} 
            F_n=Interleave(F^h_n,\ F^d_n)\ \in \ {\mathbb{R}}^{b\ \times \ h_n\times \ w_n\times \ 2d_{hiera}} 
            \end{equation} 

            This shared subspace projection offers three major benefits: (1) it reduces redundant feature overlap by enforcing a compact bottleneck; (2) it enhances robustness to domain shifts by aligning features from different pretraining distributions; and (3) it retains critical spatial structures through skip connections, improving localization fidelity.
            \subsubsection{Multi-Task Decoder}
            The Decoder consists of $\ C_s\ +\ 2C_t\ $ heads: one per source domain class and two for per target domain class (segmentation and pseudo-labeling). Each head (expect for pseudo-labeling) follows the hierarchical Decoder from \cite{zhang2025adapting}, with convolutional channels of 256, 128, and 64. For scale n, the Decoder upsamples the coarse feature map $F_{n+1}$ via transposed convolution, concatenates it with $F_n$, and applies a convolutional block:

            \begin{equation} \label{GrindEQ__4_} 
            F'_n=ConvBlock(Concat(Up(F_{n+1}),\ F_n)) 
            \end{equation} 

            where $Up(\textrm{·})\ $is transposed convolution; $Concat(\textrm{·})$ is channel-wise concatenation; $ConvBlock(\textrm{·})$ includes convolution, batch normalization, and ReLU. 

            The final head same as $ConvBlock(\textrm{·})$, and adds a $1\times 1\ $convolution to outputs logits $S\in {\mathbb{R}}^{C\ \times \ H\ \times \ W}$, where $C=2C_t$ for binary segmentation. 

            For target domain images, we introduce a lightweight pseudo-labeling head to generate pseudo-labels ${\hat{S}}_t\in {\left\{0,\ 1\right\}}^{H\ \times \ W}$ for unlabeled target domain images. It consists of three consecutive 1$\mathrm{\times}$1 convolutional layers that progressively reduce the channel dimensions from 2048 to 64, then 32, and finally to 2 channels corresponding to normal and anomalous classes, with ReLU activations after each convolution.
            
            The pseudo-labeling head outputs logits ${\mathrm{L}}_t\in R^{(2\ \times \ H\ \times \ W)}$. Softmax probabilities are computed in \textbf{Equation \ref{GrindEQ__5_}} and pixels with ${max}_cp_t(h,w,c)>\theta =0.7$ are assigned pseudo-labels in \textbf{Equation \ref{GrindEQ__6_}}.

            \begin{equation} \label{GrindEQ__5_} 
            p_t(h,w,c)=\frac{\mathrm{exp}(L_t(h,w,c))}{\sum^1_{c'=0}{\ }\mathrm{exp}(L_t(h,w,c'))}
            \end{equation} 

            \begin{equation} \label{GrindEQ__6_} 
            {\hat{S}}_t\left(h,w\right)={\mathrm{arg} \mathop{max}_{c}\ }\ p_t\left(h,w,c\right)\mathrm{\ }\mathrm{\ (}\mathrm{if\ }\mathop{max}_{c}\ p_t\left(h,w,c\right)>\theta ,\ \mathrm{else\ ignore})
            \end{equation} 

            For robustness, we apply augmentations (flipping, rotation, scaling) to target images${\mathrm{\ I}}_{\mathrm{t}}$, generating augmented images (after restore) ${\mathrm{I}}^{\mathrm{'}}_t$. The segmentation head predicts ${\mathrm{S}}^{\mathrm{'}}_t$, and consistency is enforced via:
    
            \begin{equation} \label{GrindEQ__7_} 
            L^{\mathrm{mse}}_t=\frac{1}{|P|}\sum_{(h,w)\in P}{\left(S'_t(h,w)-{\hat{S}}_t(h,w)\right)}^2 
            \end{equation} 
            
            where $P\ $is the set of pixels with valid pseudo-labels.

            The model is pre-trained on $D_s$ and ${\mathcal{D}}_t$ for $T$ epochs, followed by joint training with pseudo-labels, iterating to refine$\ {\hat{S}}_t$.
            
            The proposed MTL Decoder offers several important benefits: (1) By assigning distinct Decoder heads to each source domain category and incorporating dedicated heads for the unlabeled target domain, the model introduces domain-specific inductive biases that prevent feature conflation and encourage disentangled representation learning; (2) The integration of pseudo-labeling heads enables progressive self-supervision in the target domain, effectively bridging the supervision gap in the absence of annotated samples; (3) The hierarchical structure of the Decoder, combined with multi-scale feature fusion, enhances pixel-level localization precision, enabling the model to identify subtle and spatially diverse defects across domains. These advantages collectively allow the MTL Decoder to facilitate robust domain adaptation and fine-grained anomaly segmentation.
        \subsubsection{Loss Function}
            The total loss combines source and target domain losses: 
            \begin{equation} \label{GrindEQ__8_} 
            L=L^{\mathrm{ce}}_s+{\lambda }_1L^{\mathrm{ce}}_t+{\lambda }_2L^{\mathrm{mse}}_t 
            \end{equation} 
            \begin{equation} \label{GrindEQ__9_} 
            L^{\mathrm{ce}}_s=-\frac{1}{N_sHW}\sum^{N_s}_{i=1}\sum_{h,w}\sum^1_{c=0}{\ }S_i(h,w,c)\mathrm{log}{\ }p_i(h,w,c) 
            \end{equation}

            where ${\lambda }_1$ and ${\lambda }_2$ are adjustment coefficients of $L^{\mathrm{ce}}_t$ and $L^{\mathrm{mse}}_t$.
            
            The total loss function is designed to jointly optimize both source domain supervision and target domain adaptation. Specifically, it combines two primary components: (1) the supervised segmentation loss on the labeled source domain, and (2) the consistency-based pseudo-labeling loss on the unlabeled target domain. The source domain loss ensures that the Decoder learns accurate semantic representations and boundary-sensitive segmentation maps using ground-truth annotations. Meanwhile, the target domain loss enforces consistency between predictions on augmented versions of the same input, using pseudo-labels filtered by confidence thresholds. This encourages the model to focus on stable and transferable visual cues.
            
            To balance these objectives, two weighting coefficients ${\lambda }_s$ and ${\lambda }_t\ $are introduced to adjust the relative importance of source and target losses. These hyperparameters play a critical role in preventing over-reliance on noisy pseudo-labels while ensuring sufficient adaptation to the target domain. By combining both supervised and self-supervised signals, the loss function enables NexViTAD to generalize effectively to unseen domains while preserving high localization fidelity.
    \subsection{Inference Phase}
        The inference phase of NexViTAD is specifically designed to operate in a fully unsupervised manner, enabling robust anomaly localization in unseen target domains without requiring labeled samples. In the inference step, NexViTAD removes the Decoder, and only the Encoder processes input images ${\mathrm{I}}_{\mathrm{t}}$. $M$ normal images are selected from${\ D}_t\ $to form a memory bank:
        
        \begin{equation}
        \mathcal{M}=\{I_m{\}}^M_{m=1} 
        \end{equation}

        Each image is encoded to obtain four layers' features:
        
        \begin{equation}
        F_m=Encoder(I_m)\in {\mathbb{R}}^{b\times h_n\times w_n\times d_{\mathrm{hiera}}}
        \end{equation}        
        
        Then, features are flattened across spatial dimensions for each layer:
        
        \begin{equation}
        Z_m=\mathrm{Flatten}(F_m)\in {\mathbb{R}}^{(h_n\cdot w_n)\times d_{\mathrm{hiera}}} 
        \end{equation}          
        
        Sinkhorn K-means \cite{zhang2023detect,cuturi2013lightspeed}\textit{ }clusters\textit{ }$\{Z_m{\}}^N_{m=1\ }$into $4K\ $prototypes $\{P_k{\}}^K_{k=1}$, where $P_k\in {\mathbb{R}}^{d_{\mathrm{hiera}}}$. For a test image ${\mathrm{I}}_t$, features ${\mathrm{F}}_t$ are flattened to ${\mathrm{Z}}_t$, and the anomaly score for each patch $(h,\ w)$ is:
        
        \begin{equation}
        a(h,w)=\mathop{min}_{k}\ \parallel Z_t(h,w)-P_k{\parallel }_2
        \end{equation}        
        
        The score map $A\in {\mathbb{R}}^{h_n\times w_n}\ $is upsampled to $H\ \times \ W\ $via bilinear interpolation and smoothed with a Gaussian filter $(\sigma =2)$:

        \begin{equation}
        A'=\mathrm{GaussianFilter}(\mathrm{Upsample}(A))
        \end{equation}             
        
\begin{figure}[H]
            \centering
            \includegraphics[width=1\linewidth]{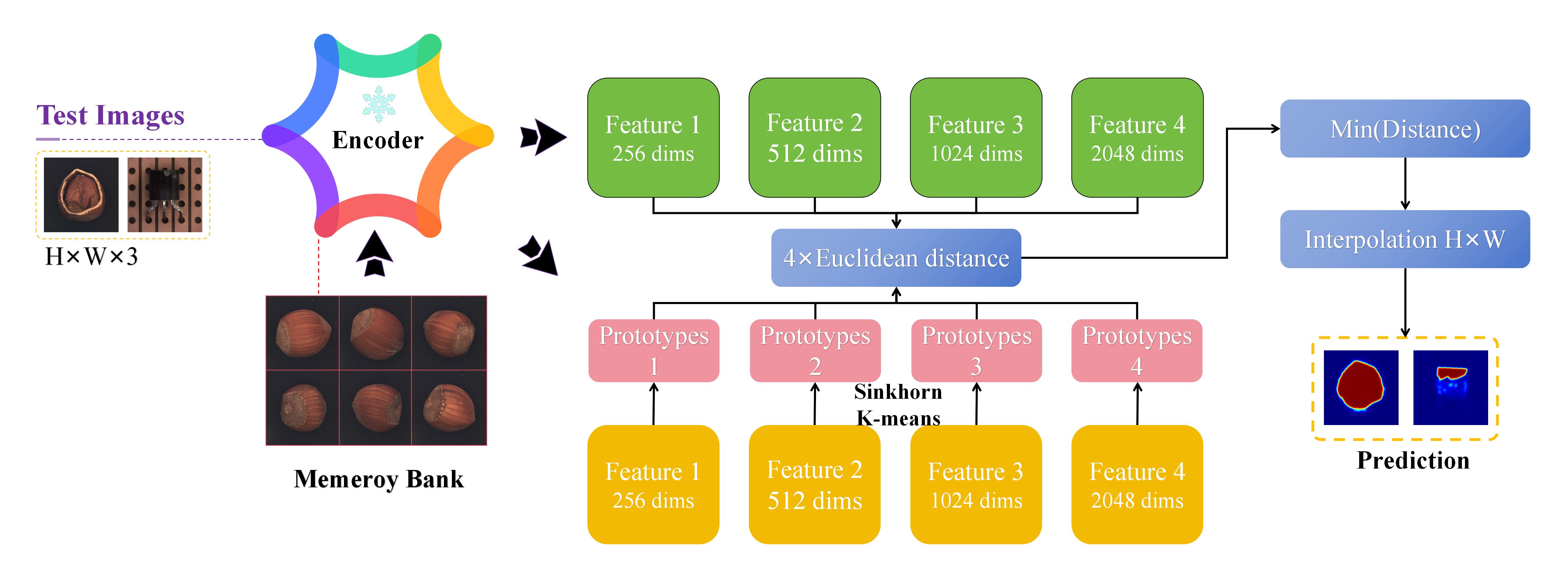}
            \caption{Example heatmap output for a target domain images, with defective regions highlighted after Gaussian smoothing using NexViTAD.}
            \label{fig:3}
        \end{figure}

        The heatmap $A'$ highlights defective regions, as shown in \textbf{Figure \ref{fig:3}}.
        
        By discarding the multi-task Decoder during testing, the framework eliminates potential overfitting to the source domain and instead relies on a memory bank of normal feature embeddings extracted from the target domain itself. These embeddings, collected from defect-free images, are clustered using the Sinkhorn K-means algorithm, which ensures structure-aware prototype generation through entropy-regularized optimal transport.
\section{Experiments}
    \subsection{Dataset and Setup}
        NexViTAD is evaluated on the MVTec AD dataset \cite{Bergmann_2019}, comprising 12 classes (containing bottle, cable, capsule, carpet, hazelnut, leather, metal-nut, pill, tile, toothbrush, transistor, wood). Images are resized to 224 $\mathrm{\times}$ 224. We use 8--11 classes as the source domain (randomly selected) and the remainder as the target domain. Each configuration is run five times with different random category selections, and average results are reported for stability. During training, only source domain labels were used, and the target domain was fully unlabeled. All experiments adopted the evaluation metrics introduced in Section 4.2. Training uses an Intel(R) Xeon(R) Gold 6230R CPU @ 2.10GHz and NVIDIA A100 GPU (40 GB) with PyTorch 2.0.0+cuda 11.8. The model is trained for $T=50$ epochs with Adam (learning rate $1\ \times \ {10}^{-4}$, linear warmup, cosine decay). Data augmentation includes flipping, rotation, scaling, and Gaussian noise (probability 0.5).
    \subsection{Evaluation Metrics}
        To comprehensively evaluate the performance of NexViTAD on target domains, we adopt a set of widely recognized quantitative metrics suitable for classification, segmentation, and anomaly detection tasks. For the target domain, where no ground truth labels are used during training, we assess anomaly detection performance using the following three metrics:
        
        Area Under the Curve (AUC): AUC reflects the model's capability to distinguish between normal and anomalous pixels across various decision thresholds. It is particularly robust under class imbalance and is computed by integrating the true positive rate (TPR) against the false positive rate (FPR).
        
        \begin{equation} \label{GrindEQ__15_} 
        \mathrm{AUC}=\int^1_0{\ }\mathrm{TPR}\left(t\right)\ d\mathrm{FPR}(t) 
        \end{equation} 
        
        where TPR is the true positive rate, and FPR is the false positive rate over anomaly score thresholds.
        
        Average Precision (AP): AP measures the area under the precision--recall curve. It emphasizes the model's ability to detect positive (defective) samples with high precision, especially under severe class imbalance. A higher AP indicates better detection precision at various recall levels.
        
        \begin{equation} \label{GrindEQ__16_} 
        AP=\sum_n{\ }\left(R_n-R_{n-1}\right)P_n\mathrm{\ } 
        \end{equation} 
        
        where ${\mathrm{P}}_n$ and ${\mathrm{R}}_n$ are precision and recall at threshold n. 
        
        Pixel-wise Region Overlap (PRO): PRO evaluates the overlap between predicted anomaly regions and the ground truth defect regions at the pixel level. It is defined as the mean IoU over all regions, and is especially informative for pixel-wise anomaly localization performance.
       
        \begin{equation} \label{GrindEQ__17_} 
        \mathrm{PRO}=\frac{1}{T}\sum^T_{t=1}{\ }\frac{|{\hat{S}}_t\cap S_t|}{|{\hat{S}}_t\cup S_t|}\  
        \end{equation} 
        
        where ${\hat{S}}_t$ is the predicted segmentation, $S_t$ is the ground truth, and $T\ $is the number of test images.
        
        These metrics together offer a comprehensive evaluation framework to analyze the performance of NexViTAD across different tasks and domain settings, ensuring reliable and reproducible results under both supervised and unsupervised conditions.
    \subsection{Target Domain Performance}
        To evaluate the cross-domain generalization ability of NexViTAD, we conduct extensive experiments under four different source/target configurations: 11/1, 10/2, 9/3, and 8/4, where the first number indicates the number of source domain classes and the second indicates the number of target domain classes. The number of clusters is 30, and the batch size of good images is 10. Each configuration is repeated 20 times with random splits, and the best values are reported.
        
        \textbf{Table \ref{tab:1}} summarizes the performance of NexViTAD on the unlabeled target domains, measured by three standard anomaly detection metrics: AUC, AP, and PRO. NexViTAD maintains a high AUC across all domain split settings, dropping only modestly from 97.53\% to 95.93\% as the number of target classes increases. This confirms the model's robustness in distinguishing normal from anomalous regions, even with minimal supervision. However, the AP score drops more significantly (from 70.36\% to 52.65\%), indicating that detection precision is affected by the growing complexity and diversity of the target domain. Despite this, the PRO metric surprisingly improves from 95.17\% to 99.09\%, suggesting that the model remains highly effective in localizing the overall shape and position of defect regions, even if some boundaries or pixel-level confidence scores are suboptimal. This inverse trend between AP and PRO indicates that NexViTAD emphasizes spatial coverage over prediction certainty when generalized to broader target distributions---an expected behavior in unsupervised scenarios where pseudo-label is noisier and prototypes less discriminative.
        
        \begin{table}[H]
            \caption{NexViTAD target domain performance under different source/target splits.}
            \centering
            \begin{tabular}{llll}
                \toprule 
                \textbf{Metrics} & \textbf{AUC (\%)} & \textbf{AP (\%)} & \textbf{PRO (\%)} \\ \hline 
                11/1 & 97.53 & 70.36 & 95.17 \\  
                10/2 & 97.19 & 65.54 & 95.00 \\  
                9/3 & 95.96 & 60.72 & 96.36 \\  
                8/4 & 95.93 & 52.65 & 99.09 \\
                \bottomrule
            \end{tabular}
            \label{tab:1}
        \end{table}
        
\begin{figure}[H]
            \centering
            \includegraphics[width=1\linewidth]{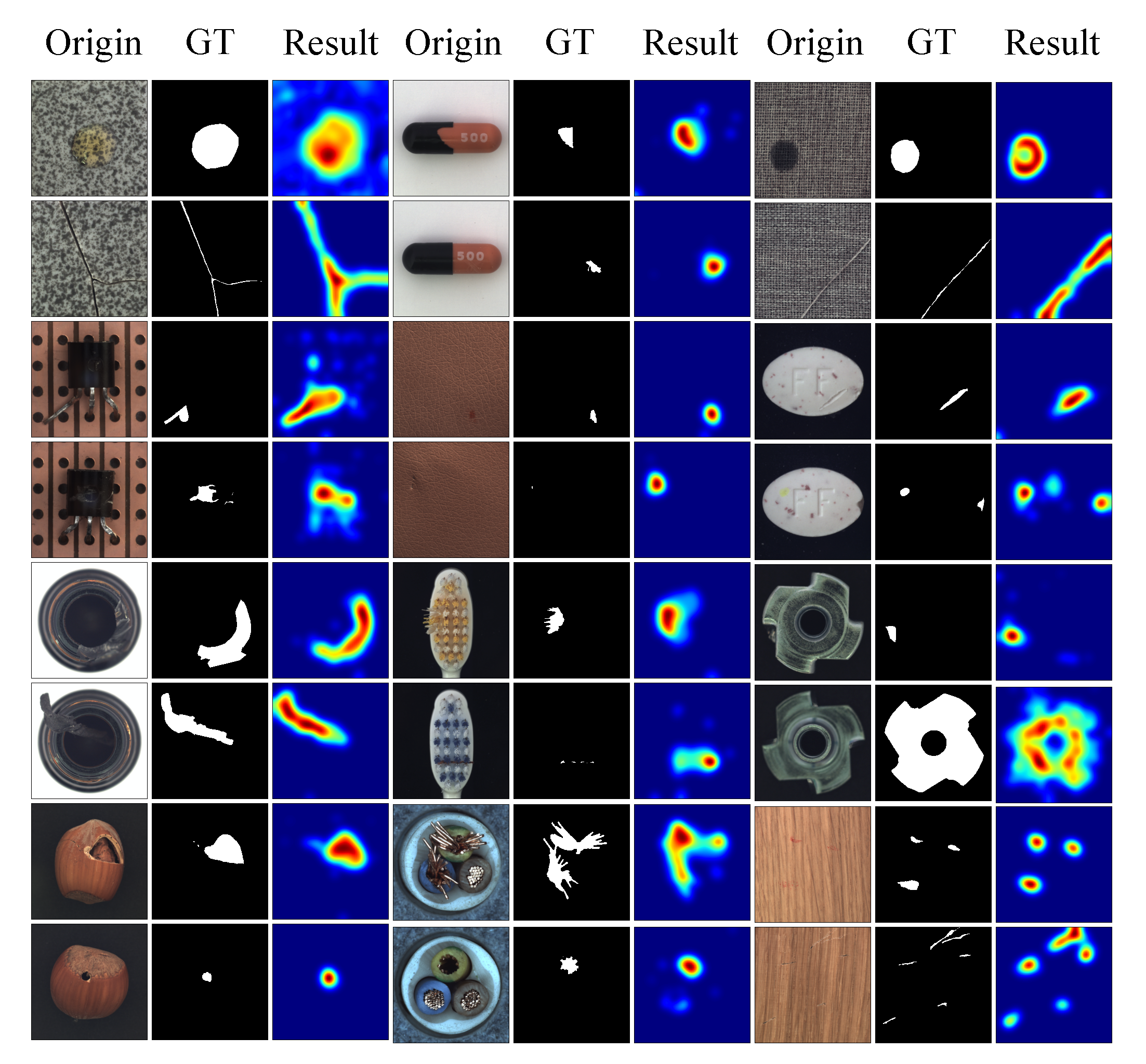}
            \caption{Qualitative defect detection results on MVTec AD target domains (11/1 split), containing original images, ground truth, and heatmaps of predicted results.}
            \label{fig:4}
        \end{figure}

        \textbf{Figure \ref{fig:4}} illustrated the defect detection results for NexViTAD on the MVTec AD dataset, specifically for the 11/1 domain split configuration. It shows original images, ground truth defect masks, and the defect detection results produced by NexViTAD across 12 different categories. From the results, it observes that NexViTAD effectively localizes defects across a wide range of categories, demonstrating strong pixel-level accuracy in detecting anomalies. For example, in the hazelnut and tile categories, where defects are subtle and the texture is complex, the model successfully highlights the defective areas without being misled by background noise. The cable and metal nut categories, which feature intricate patterns and high contrast, further validate the model's robustness in handling challenging defect structures. The capsule and leather categories illustrate model's ability to detect defects even in highly uniform textures, where traditional methods may struggle due to lack of variation. Furthermore, the results show clear delineation of the defect regions, confirming the model's capability to produce precise anomaly maps even in the absence of labeled data for the target domain. These qualitative results further solidify NexViTAD's superior performance in cross-domain defect detection, with exceptional generalization across diverse defect types and domains. For additional reference, please see \textbf{Figures \ref{fig:S1}--\ref{fig:S3}} in the Appendix for the 10/2, 9/3, and 8/4 domain splits, where the detection performance is also notably strong, further supporting the effectiveness of NexViTAD across different configurations.
        
        After analyzing the performance trend across domain splits, we now benchmark NexViTAD against existing cross-domain anomaly detection methods, including XDNet, and CAI, data sources from \cite{Lee_2023,Shi_2025}. NexViTAD sets a new state-of-the-art on the target domain, outperforming all existing cross-domain anomaly detection methods, including XDNet and both CAI variants. Specifically, NexViTAD attains an AUC of 97.5 \%, an AP of 70.4 \% and a PRO of 95.2 \%, representing absolute gains of 0.9 \%, 5.3 \% and 2.8 \%, respectively, over the XDNet. These improvements are particularly striking in the AP and PRO metrics, underscoring NexViTAD's enhanced ability to localize anomalies at the pixel level and to align anomalous regions across domains without access to target-domain labels. Such gains demonstrate that our integration of domain-adaptive vision transformers with anomaly-aware decoding not only elevates classification robustness (as reflected in AUC) but also delivers substantial precision in delineating defect boundaries and capturing spatial consistency (as reflected in AP and PRO) compared to prior approaches. 

        \begin{table}[H]
         \caption{Target domain performance comparison.}
          \centering
            \begin{tabular}{llll}
                \toprule
                \textbf{Method} & \textbf{AUC (\%)} & \textbf{AP (\%)} & \textbf{PRO (\%)} \\ \hline 
                CAI-N \cite{Shi_2025} & 95.6 & 62.6 & 87.5 \\ 
                CAI-M \cite{Shi_2025} & 94.7 & 61.6 & 86.6 \\ 
                XDNet \cite{Lee_2023} & 96.6 & 65.1 & 92.4 \\
                NexViTAD & 97.5 & 70.4 & 95.2 \\ 
                \bottomrule
          \end{tabular}
          \label{tab:table}
        \end{table}
    \subsection{Ablation Study}
        To thoroughly evaluate the impact of key components in the NexViTAD framework, we conduct a study across four critical factors: (1) Decoder configuration during inference, (2) the number of clusters in the inference phase, (3) the use of pseudo-labeling during training, and (4) the application of multi-task learning (MTL) during training. Each experiment is conducted across multiple domain splits: 11/1, 10/2, 9/3, and 8/4, to ensure the robustness of the model under varying conditions. The results from these experiments are presented and discussed to highlight the importance of each component in achieving optimal performance.
        \subsubsection{Impact of Decoder Configuration and Pseudo-labeling Head}
            In the NexViTAD, the original Decoder is used exactly as trained during the training phase, while the memory bank Decoder leverages a memory bank of normal sample features for clustering and anomaly score calculation during inference. This enables the model to more effectively detect anomalies in target domain images without needing to rely on a fixed, pre-trained Decoder. To evaluate its effect, we compare the performance of two Decoder setups during inference: (1) using the Decoder trained, and (2) using the memory bank-based Decoder, which leverages the normal sample memory bank for anomaly detection. This ablation allows us to analyze whether the use of memory bank-based inference enhances defect detection performance. The results of heatmap images are shown in \textbf{Figures \ref{fig:s4}--\ref{fig:s7}} in the appendix B for the 10/2, 9/3, and 8/4 domain splits.
            
\begin{figure}[H]
            \centering
            \includegraphics[width=1\linewidth]{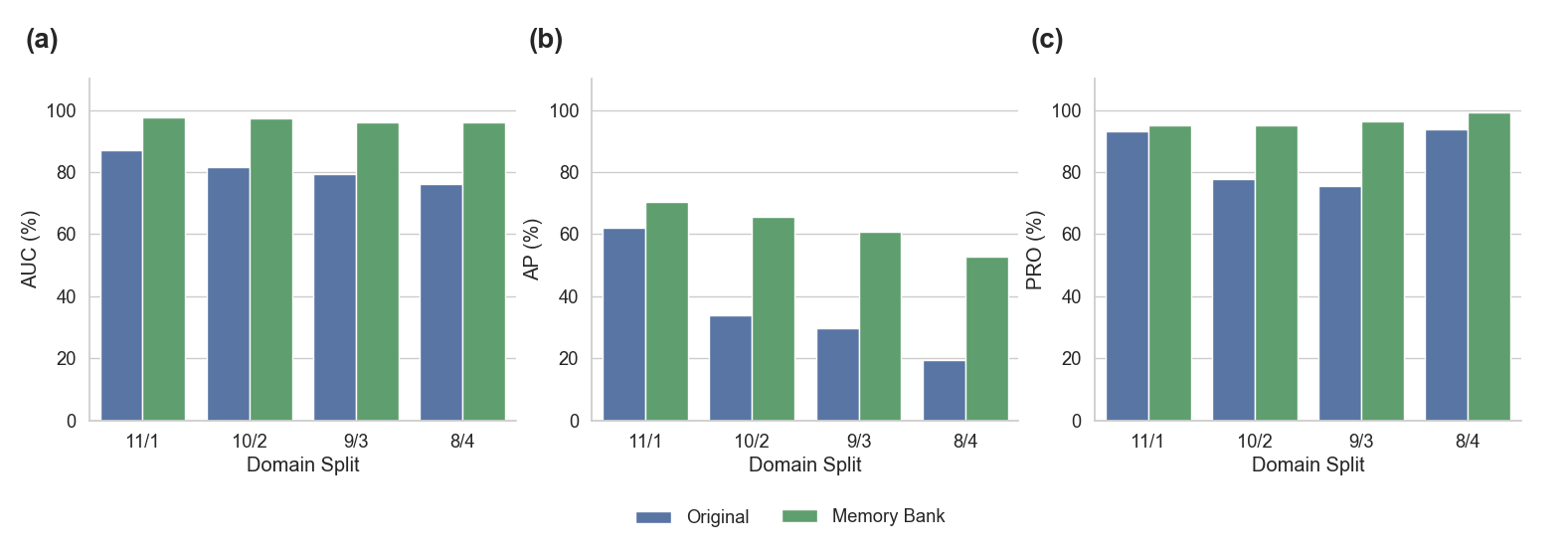}
            \caption{Impact of Decoder configuration during inference, showing the performance comparison of the original Decoder and memory bank Decoder in terms of (a) AUC, (b) AP, and (c) PRO across the 11/1, 10/2, 9/3, and 8/4 domain splits.}
            \label{fig:5}
            \end{figure}

            The metrics of the comparison between the original Decoder and memory bank Decoder are shown in Figure \ref{fig:5}. As expected, the memory bank Decoder consistently outperforms the original Decoder across all four domain splits. On the 11/1 split, it improves AUC from 87.13\% to 97.53\%, AP from 62.06\% to 70.36\%, and PRO from 93.20\% to 95.17\%. In the 10/2 split, AUC rises from 81.63\% to 97.19\%, AP nearly doubles from 33.74\% to 65.54\%, and PRO improves from 77.64\% to 95.00\%. In the 9/3 split, AUC rises from 79.43\% to 95.96\%, AP from 29.55\% to 60.72\%, and PRO from 75.38\% to 96.36\%. Most notably, in the most challenging 8/4 case, AUC improves from 76.06\% to 95.93\%, AP from 19.37\% to 52.65\%, and PRO from 93.93\% to 99.09\%. These consistent improvements stem from the memory bank's ability to dynamically represent normal feature distributions at test time without relying on source domain supervision. By clustering normal features using Sinkhorn K-means, the model establishes structure-aware prototypes that capture the underlying distribution of target domain data. This unsupervised matching mechanism allows NexViTAD to localize anomalies more accurately, especially when pseudo-labels are noisy or unavailable, and significantly mitigates the negative transfer risks associated with fixed Decoder inference.
            
            Therefore, the memory bank--based Decoder enables NexViTAD to adapt better to target domains by leveraging prototype-based similarity rather than relying on learned mappings from the source domain.

\begin{figure}[H]
            \centering
            \includegraphics[width=1\linewidth]{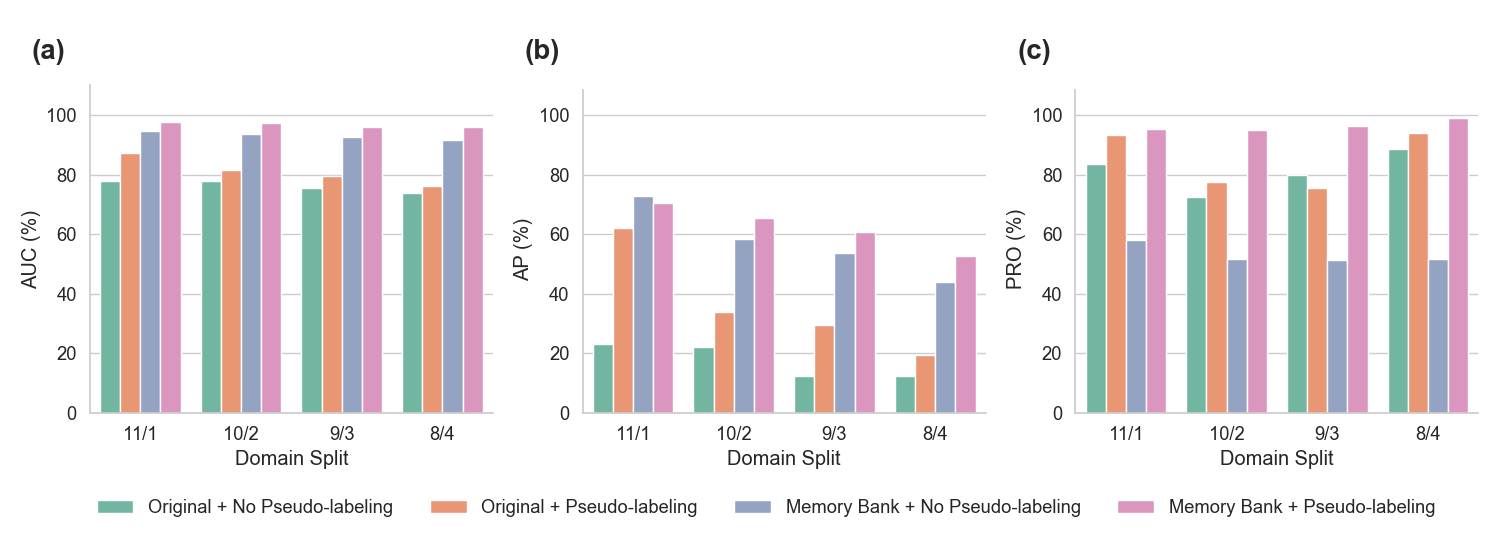}
            \caption{Effect of pseudo-labeling and Decoder type during training, showing the performance comparison with or without pseudo-labeling head in terms of (a) AUC, (b) AP, and (c) PRO across the 11/1, 10/2, 9/3, and 8/4 domain splits.}
            \label{fig:6}
            \end{figure}

            To examine the effectiveness of pseudo-labeling head, we then compare the full NexViTAD framework with pseudo-labeling against a variant where the pseudo-labeling head is removed. As shown in \textbf{Figure \ref{fig:6}}, both the use of pseudo-labeling and the choice of Decoder architecture have a significant impact on the model's performance across all domain splits. We observe consistent and interpretable trends in AUC, AP, and PRO under eight experimental configurations: combining two Decoder types (Original vs. Memory Bank) and two supervision modes (with vs. without pseudo-labeling head). For the original Decoder, the introduction of pseudo-labeling consistently improves anomaly detection metrics. In the 11/1 split, AUC increases from 77.90\% to 87.13\%, AP from 23.23\% to 62.06\%, and PRO from 83.54\% to 93.20\%. Similarly, in the 10/2 split, AUC rises from 77.68\% to 81.63\%, AP from 22.28\% to 33.74\%, and PRO from 72.49\% to 77.64\%. In the 9/3, gains from 75.37\% to 79.43\% (AUC), 12.28\% to 29.55\% (AP), and 79.99\% to 75.38\% (PRO), with the slight drop in PRO attributed to Decoder rigidity. In the 8/4 split, pseudo-labeling improves AUC from 73.65\% to 76.06\%, AP from 12.39\% to 19.37\%, and PRO from 88.45\% to 93.93\%. As using the Memory Bank Decoder, the improvements are even more substantial. With pseudo-labeling head, AUC rises from 94.64\% to 97.53\% in the 11/1 split, AP from 72.72\% to 70.36\% (a slight dip due to over-saturation), and PRO from 57.89\% to 95.17\%. In the 10/2, AUC improves from 93.70\% to 97.19\%, AP from 58.21\% to 65.54\%, and PRO from 51.67\% to 95.00\%. In the 9/3, AUC grows from 92.64\% to 95.96\%, AP from 53.53\% to 60.72\%, and PRO from 51.24\% to 96.36\%. In the 8/4, AUC improves from 91.75\% to 95.93\%, AP from 43.92\% to 52.65\%, and PRO from 51.50\% to 99.09\%. The results of heatmap images could be found in \textbf{Figures \ref{fig:s4}--\ref{fig:s7}}.
            
            These results reveal two critical insights. First, pseudo-labeling head plays a central role in enabling unsupervised target domain adaptation: it provides weak but progressively refined supervision to the Decoder, allowing it to adapt its feature understanding of the target domain even in the absence of ground-truth labels. The confidence filtering and perturbation consistency constraints jointly ensure that the model focuses on stable and meaningful structural cues, avoiding noise amplification. Second, the memory bank, by relying on prototype-based matching rather than Decoder prediction, demonstrates superior robustness to domain shifts. It encodes normal feature distributions in a structure-aware manner using Sinkhorn K-means, enabling precise anomaly localization regardless of the Decoder's training state. 
            
            Therefore, these findings confirm that pseudo-labeling and memory bank inference are complementary: pseudo-labels improve the feature representations and domain alignment during training, while the memory bank Decoder offers a reliable and flexible inference strategy during testing. Their combination yields the strongest performance across all metrics and settings, highlighting NexViTAD's capability to generalize under minimal supervision in real-world defect detection scenarios. 
            
            \subsubsection{Impact of MTL in Training}
                To assess the impact of MTL in Decoder during training process, two training configurations under consistent inference conditions using the memory bank-based Decoder are compared. The first configuration adopts a multi-Decoder MTL framework, where each source domain has an individual Decoder head, and two additional Decoders is assigned to pseudo-labeled target data. The second configuration employs a single shared Decoder for both source domain data, and two additional Decoders to target data, representing a simplified variant without multi-task supervision. 
                
\begin{figure}[H]
                \centering
                \includegraphics[width=1\linewidth]{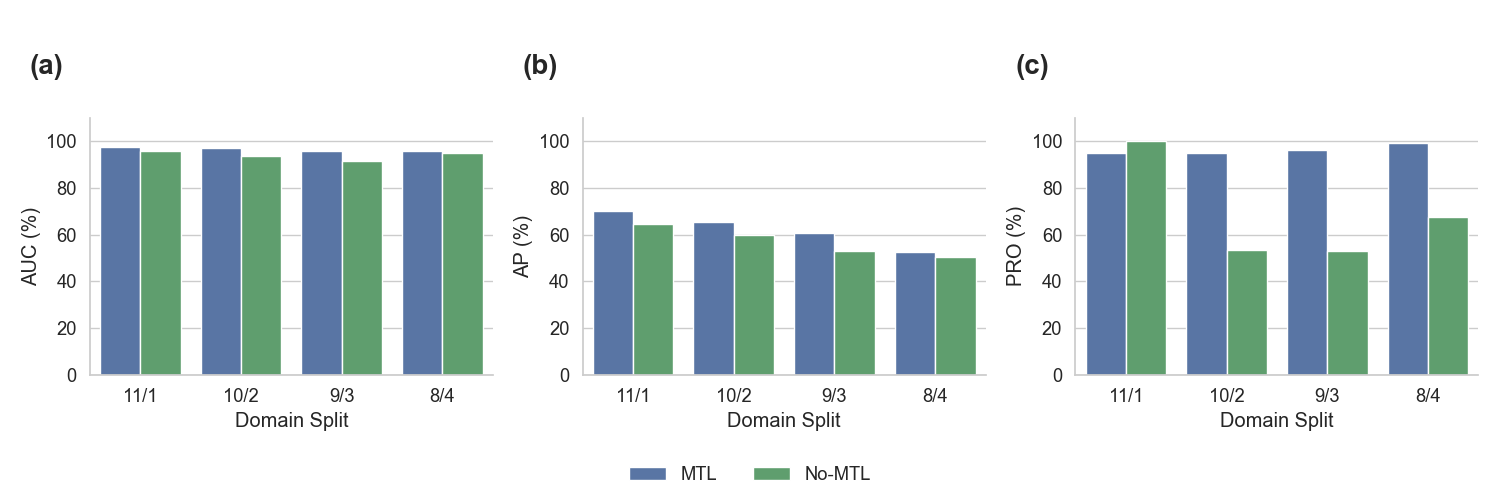}
                \caption{Effect of MTL during training: (a) AUC, (b) AP, and (c) PRO across the 11/1, 10/2, 9/3, and 8/4 domain splits.}
                \label{fig:7}
                \end{figure}

                As shown in \textbf{Figure \ref{fig:7}}, the MTL-based configuration consistently achieves superior performance. For the 11/1 split, AUC increases from 95.93\% to 97.53\%, AP rises from 64.47\% to 70.36\%, and PRO slightly decreases from 99.92\% to 95.17\%, likely due to sharper localization boundaries. In the 10/2 split, the improvements are more substantial: AUC increases by 3.30\%, AP improves by 5.64\%, and PRO significantly improves by 41.47\%. For the 9/3 split, AUC and AP improve by 4.27\% and 7.59\% respectively, while PRO sees a dramatic gain of 43.12\%. Similarly, in the 8/4 split, MTL leads to an AUC gain of 1.14\%, AP increase of 2.27\%, and PRO improvement of 31.63\%. The results of heatmap images could be found in \textbf{Figures \ref{fig:s8}--\ref{fig:s11}}. The superior performance of the MTL-based configuration stems from its ability to introduce domain-specific inductive biases during training. By assigning each source domain its own Decoder, the model can learn fine-grained, domain-aware representations without conflating domain-specific characteristics. Meanwhile, the pseudo-labeled Decoder encourages alignment with target-domain semantics. This structure facilitates the learning of disentangled yet complementary features, which are subsequently aggregated in the unified memory bank Decoder during inference. Overall, the results validate that incorporating MTL is a highly effective strategy for improving both the discriminative power and localization accuracy of cross-domain anomaly detection models.
                
            \subsubsection{Impact of Cluster Number in Inference}
\begin{figure}[H]
                \centering
                \includegraphics[width=1\linewidth]{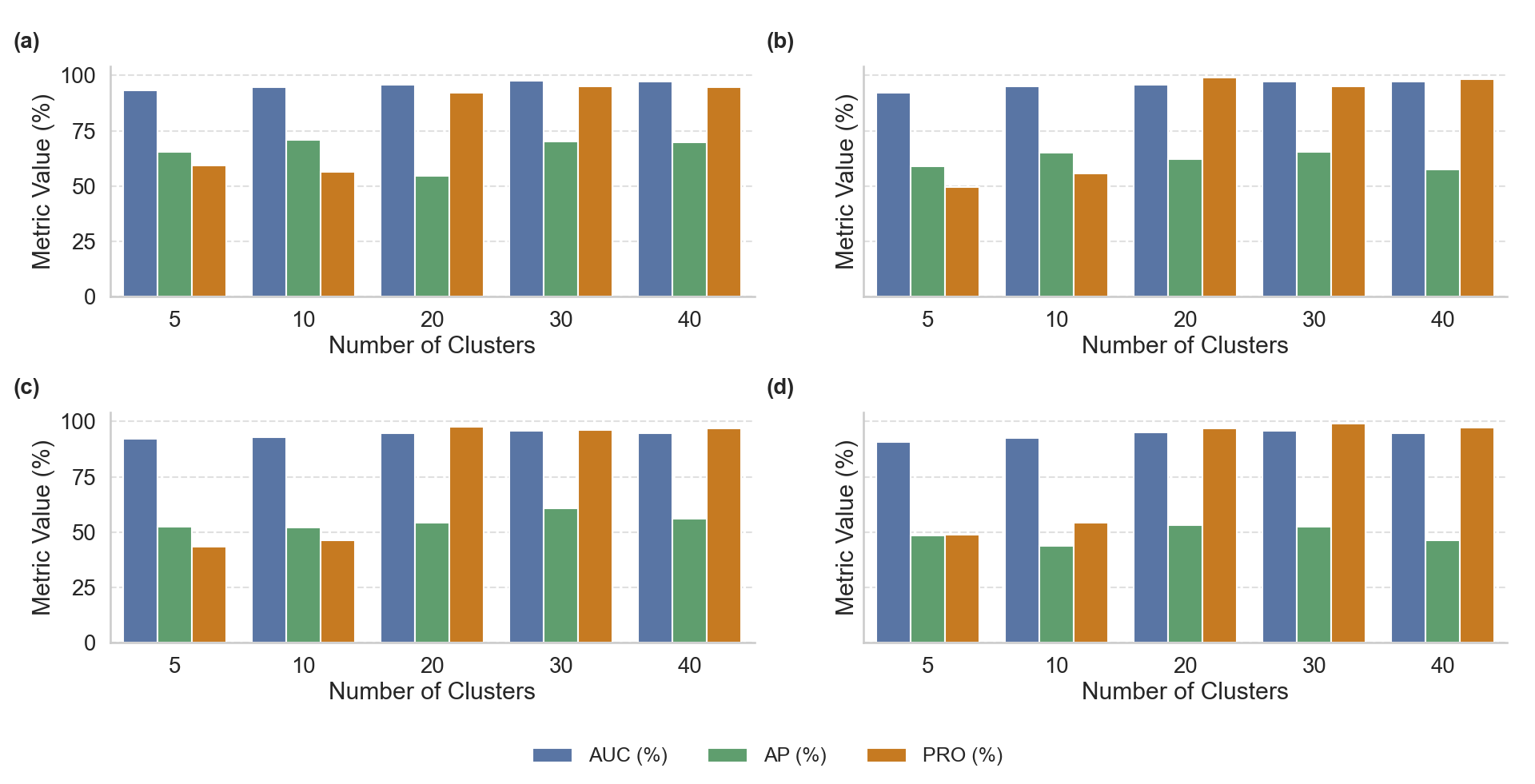}
                \caption{Effect of cluster number on AUC, AP, and PRO across splits: (a) 11/1, (b) 10/2, (c) 9/3 and (c) 8/4.}
                \label{fig:8}
                \end{figure}

                To assess the effect of clustering granularity during inference, we investigate how varying the number of prototypes (clusters) affects anomaly detection performance under the full configuration (MTL + pseudo-labeling head + memory bank), across all four domain splits (11/1, 10/2, 9/3, 8/4). The selected cluster numbers are: 5, 10, 20, 30, and 40. The corresponding results are summarized in \textbf{Figure \ref{fig:8}} and the heatmap images could be found in \textbf{Figures \ref{fig:s12}--\ref{fig:s15}}.
                
                As only 5 clusters, the model exhibits limited performance across all splits, with AUC ranging from 90.67\% (8/4) to 93.42\% (11/1), and AP/PRO scores consistently lower (e.g., 48.71\% and 48.82\% on 8/4). This performance bottleneck can be attributed to the insufficient representational capacity of a small prototype set, which fails to capture the nuanced diversity of visual concepts across object parts and defect patterns. The coarse representation leads to blurred anomaly localization and lower confidence in pixel-wise classification. Increasing the number of clusters to 10 yields marginal improvements in AUC and AP, notably on 11/1 (AUC: 94.78\%, AP: 70.90\%) and 10/2 (AUC: 95.15\%, AP: 65.10\%). However, the PRO values remain low in this setting (e.g., 56.40\% on 11/1), indicating that although the anomaly scores may be improved at the pixel level, the region-level consistency of localization is still suboptimal. This highlights that mere expansion of the prototype space does not yet suffice to model structural coherence. At 20 clusters, the model begins to show significantly better performance across splits, particularly in PRO. All domain splits exceed 94.7\% AUC and reach above 96.9\% PRO (e.g., 99.05\% on 10/2), demonstrating enhanced ability to capture both global semantics and spatial structures. However, AP on some splits (e.g., 11/1: 54.69\%, 9/3: 54.21\%) slightly decreases, suggesting that the increased number of prototypes may introduce redundant or overlapping features, leading to reduced pixel-wise precision.
                
                The best trade-off is achieved as 30 clusters. This configuration consistently delivers optimal performance across all three metrics: AUC (95.93\%--97.53\%), AP (52.65\%--70.36\%), and PRO (95.00\%--99.09\%). On the 11/1 split, for instance, AUC reaches 97.53\%, AP achieves 70.36\%, and PRO hits 95.17\%. These gains are underpinned by the semantic balance between diversity and compactness in the prototype space. With 30 clusters, the model is able to differentiate subtle variations in normal features while avoiding over-fragmentation, thus enhancing the robustness of anomaly highlighting. Further increasing the cluster number to 40 yields saturated or slightly degraded results. While AUC and PRO remain high (e.g., 97.36\% and 94.64\% on 11/1), AP exhibits a notable decline in some splits, such as from 70.36\% to 69.86\% on 11/1 and from 65.54\% to 57.40\% on 10/2. This reflects the curse of over-partitioning, where excessive clusters may cause feature dispersion, diluting meaningful groupings and introducing noise sensitivity during similarity aggregation. 
                
                In essence, cluster number controls the granularity of semantic representation during inference. Too few clusters underfit the diversity of visual concepts, while too many clusters lead to over-fragmentation and noisy attention. Setting the number of clusters to 30 achieves the optimal balance between expressiveness and stability, yielding superior anomaly localization and discrimination across all domain scenarios.
                
            \subsubsection{Comparison of Inference Time}
\begin{figure}[H]
                \centering
                \includegraphics[width=0.5\linewidth]{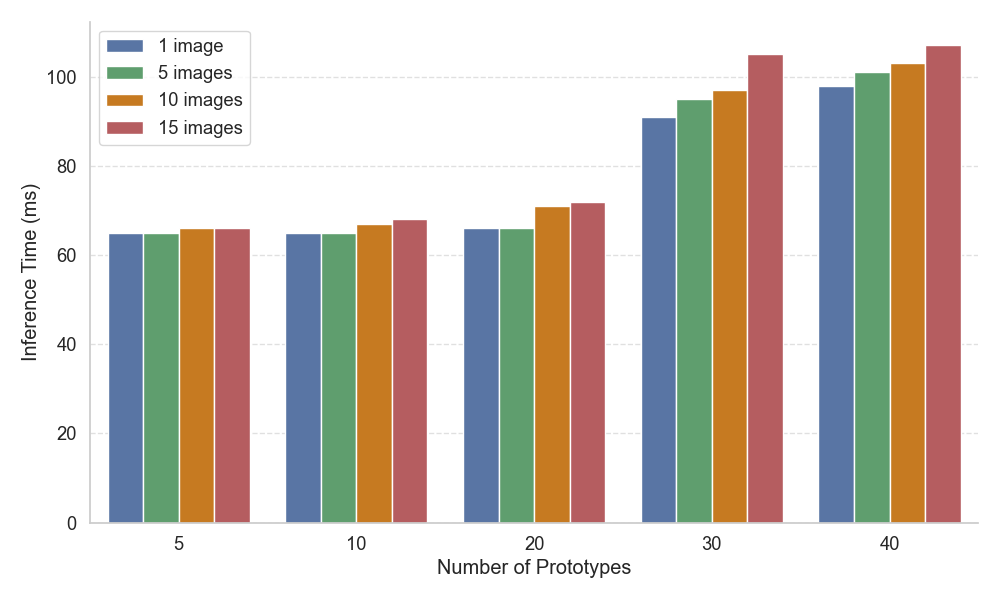}
                \caption{Inference time (ms) vs. number of prototypes for different batch sizes.}
                \label{fig:9}
                \end{figure}

                The inference time analysis presented in \textbf{Figure \ref{fig:9}} comprehensively evaluates the impact of varying prototype quantities and image batch sizes on the overall computational latency, including 35 ms feature extraction overhead. The results demonstrate a clear linear relationship between the number of prototypes and total inference time, which becomes particularly pronounced when the prototype count exceeds 20. Specifically, as the number of prototypes increases from 5 to 20, the inference time grows marginally across all batch sizes from 65 ms to 72 ms indicating that the computational overhead introduced by prototype matching remains manageable at low to moderate prototype scales. However, once the number of prototypes increases to 30 and 40, the inference time escalates more noticeably. At a batch size of 15, the inference time increases from 72 ms (20 prototypes) to 105 ms (30 prototypes) and further to 107 ms (40 prototypes), suggesting a near-linear increase in computational cost due to repeated feature-prototype comparisons.
                
                Moreover, increasing the batch size from 1 to 15 consistently results in slightly higher inference times, yet the increments remain relatively small typically under 10 ms per batch size step. This moderate growth reflects the model's efficient batch processing capability and its scalability under real-world deployment settings. These comparisons grow linearly with the number of prototypes and accumulate over the batch dimension. Despite this, the model exhibits consistent processing performance under small to medium batch sizes, maintaining total inference time below 100 ms for up to 30 prototypes and 10 images, making it suitable for near-real-time industrial applications.
                
    \section{Conclusion}
        A novel framework, NexViTAD, was proposed for few-shot unsupervised cross-domain defect detection and rigorously evaluated on the MVTec AD dataset. The framework achieved state-of-the-art performance, with an AUC of 97.5\%, AP of 70.4\%, and PRO of 95.2\% on target domains, outperforming baseline methods (MTL, pseudo-labels, 11/1 split, 30 prototypes, and 10 defect-free images). These results underscore NexViTAD's superior generalization across diverse industrial scenarios, evidenced by consistent performance across domain splits. The integration of the hierarchical vision transformer Hiera for multi-scale feature extraction, combined with DINOv2 features for enhanced semantic representation, enables robust adaptation to varied defect types and domains. The multi-task Decoder, incorporating task-specific heads and a confidence-based pseudo-labeling strategy, facilitates effective learning on unlabeled target domains, while the inference pipeline, utilizing Sinkhorn K-means clustering with a memory bank of normal samples, ensures precise pixel-level anomaly localization.
        
        Despite these advancements, limitations were identified. The computational complexity, with inference times may hinder real-time applications. Performance degradation was observed under extreme domain shifts, particularly in the 8/4 split, reflecting challenges with highly diverse defect types. Furthermore, reliance on a memory bank of normal samples may constrain applicability in scenarios where normal samples are scarce. Future research should focus on reducing computational overhead through optimized architectures, improving domain adaptation for larger domain gaps, and exploring applications in fields like medical imaging or autonomous systems. NexViTAD establishes a robust foundation for advancing cross-domain defect detection, demonstrating significant potential for industrial quality assurance.

    \section*{Acknowledgments}
        This work was supported by Liaoning Provincial Education Department Fund Project: Digital twin-driven global resilience assessment model and dynamic simulation research, No.202464252-1.

\bibliographystyle{unsrt}  
\bibliography{references}  
\newpage
\section{Appendix}
\setcounter{figure}{0}
\renewcommand*{\thefigure}{S\arabic{figure}}
\raggedbottom
            \begin{figure}[H]
                \centering
                \includegraphics[width=0.75\linewidth]{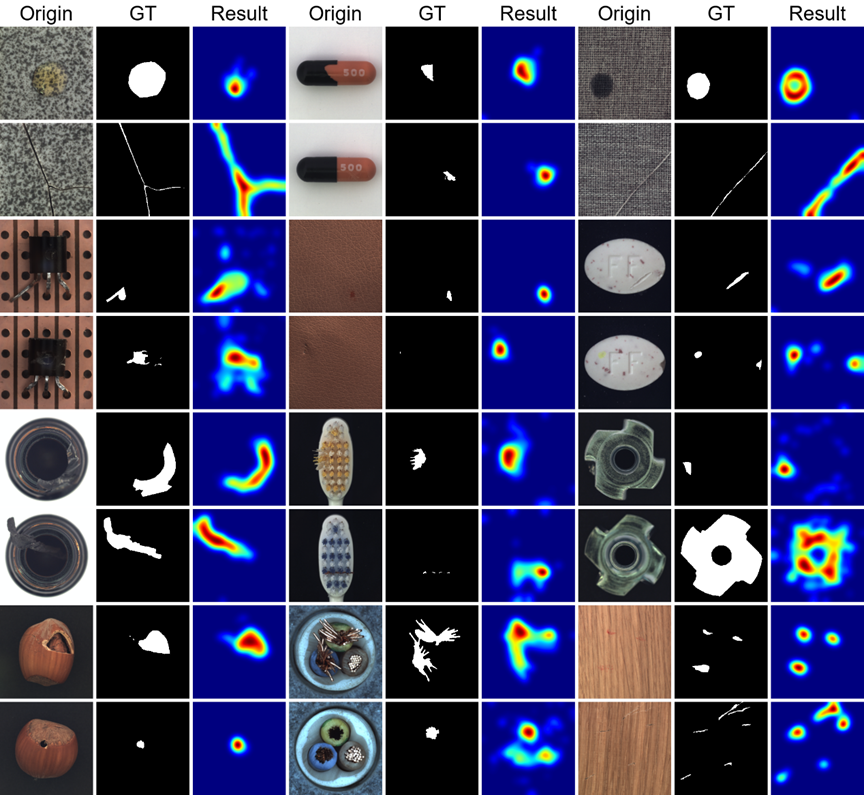}
                \caption{Qualitative defect detection results on MVTec AD target domains (8/4 Split), containing original images, ground truth, and heatmaps of predicted results.}
                \label{fig:S1}
            \end{figure}
            \begin{figure}[H]
                \centering
                \includegraphics[width=0.75\linewidth]{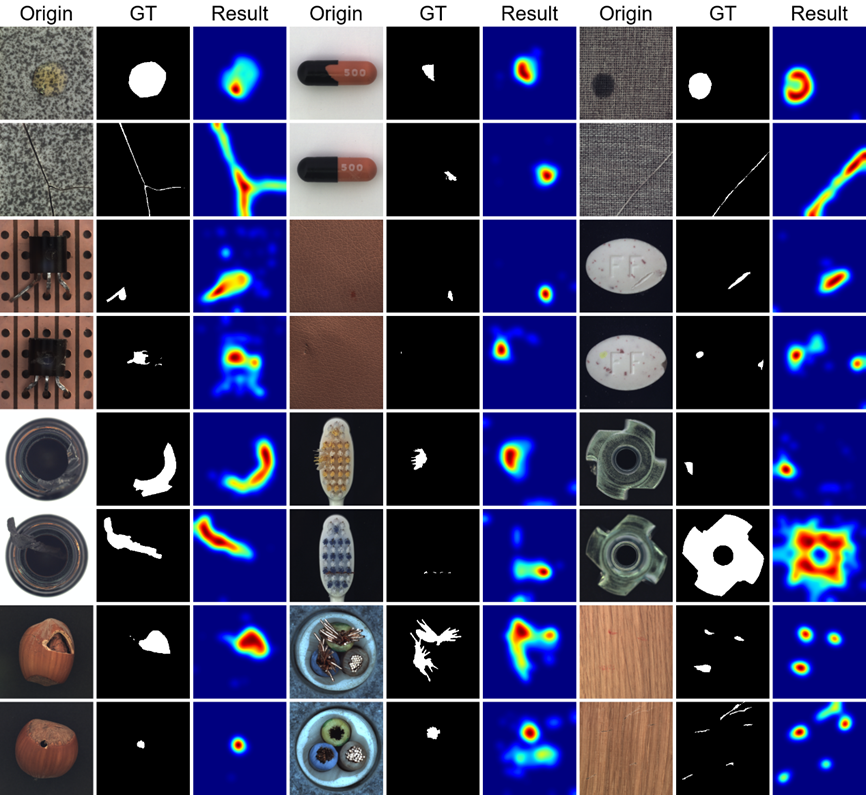}
                \caption{Qualitative defect detection results on MVTec AD target domains (9/3 Split), containing original images, ground truth, and heatmaps of predicted results.}
                \label{fig:S2}
            \end{figure}
             \begin{figure}[H]
                 \centering
                 \includegraphics[width=0.75\linewidth]{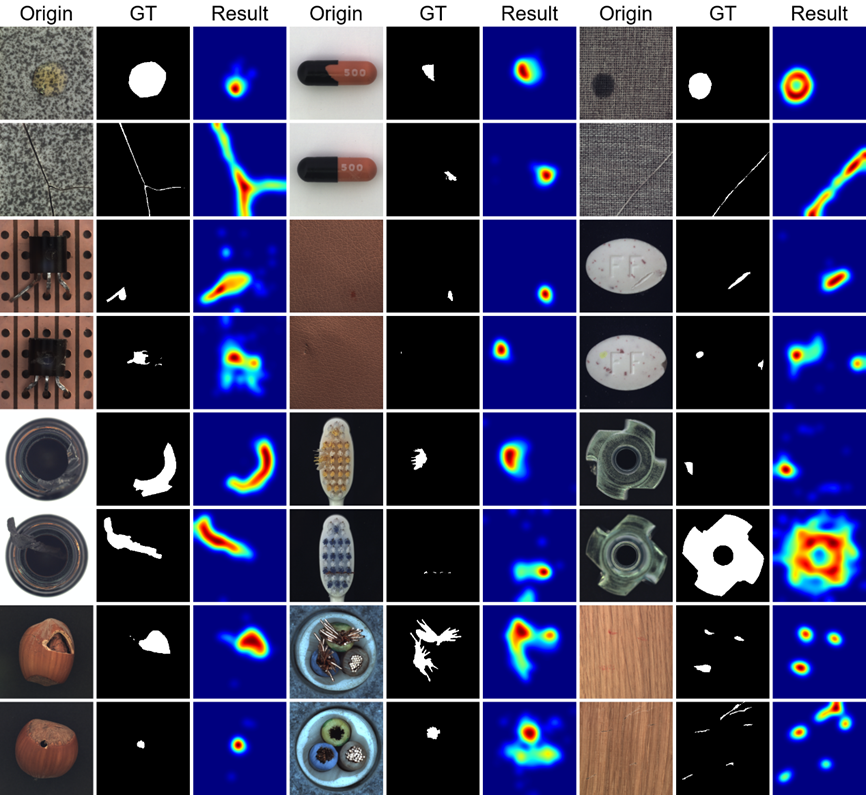}
                 \caption{Qualitative defect detection results on MVTec AD target domains (10/2 Split), containing original images, ground truth, and heatmaps of predicted results.}
                 \label{fig:S3}
             \end{figure}
    \begin{figure}
        \centering
        \includegraphics[width=0.75\linewidth]{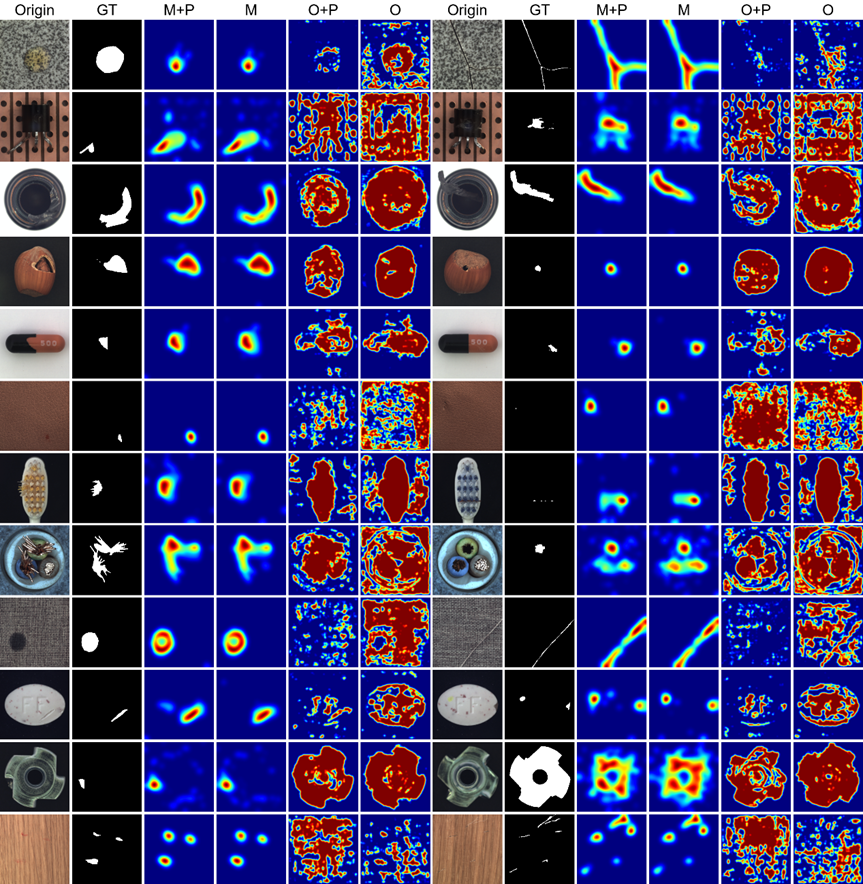}
        \caption{Impact of Decoder configuration and pseudo-labeling head during inference, results on MVTec AD Target Domains (8/4 split): M = memory bank; P = pseudo-labeling; O = original Decoder inference.}
        \label{fig:s4}
    \end{figure}
    \begin{figure}
        \centering
        \includegraphics[width=0.75\linewidth]{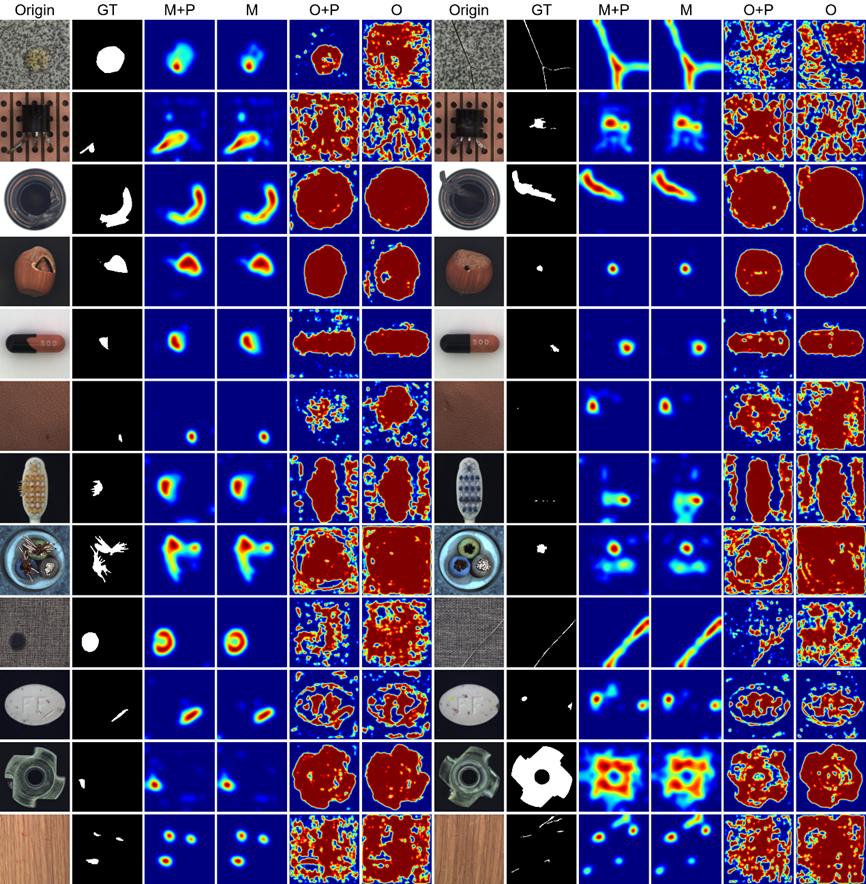}
        \caption{Impact of Decoder configuration and pseudo-labeling head during inference, results on MVTec AD Target Domains (9/3 split): M = memory bank; P = pseudo-labeling; O = original Decoder inference.}
        \label{fig:s5}
    \end{figure}
    \begin{figure}
        \centering
        \includegraphics[width=0.75\linewidth]{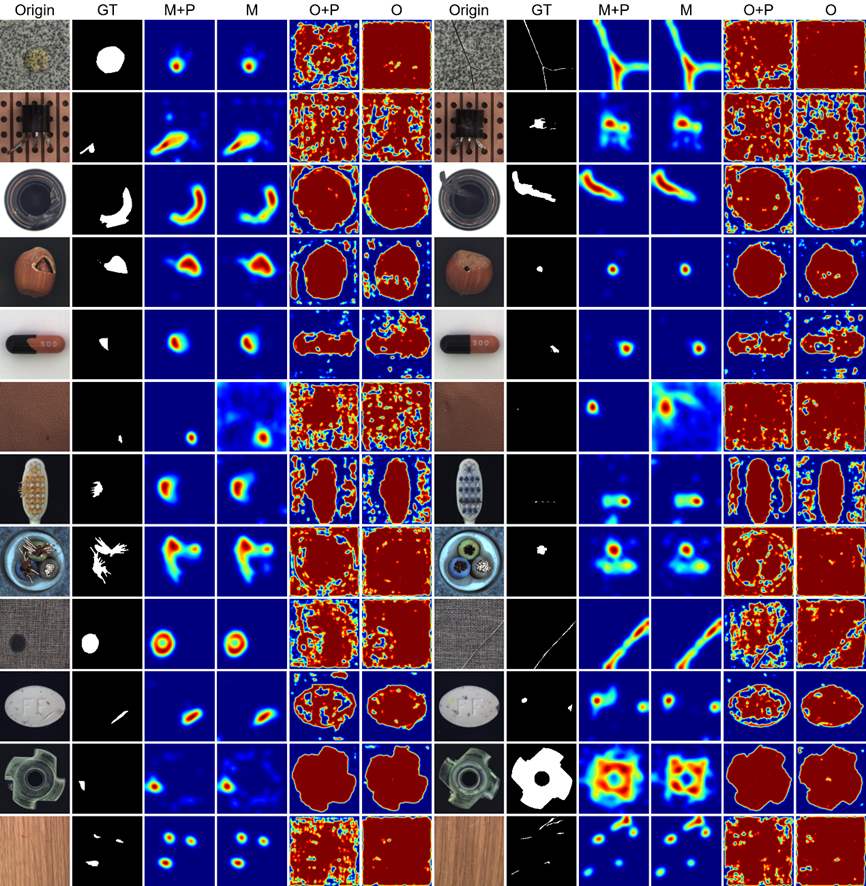}
        \caption{Impact of Decoder configuration and pseudo-labeling head during inference, results on MVTec AD Target Domains (10/2 split): M = memory bank; P = pseudo-labeling; O = original Decoder inference.}
        \label{fig:s6}
    \end{figure}
    \begin{figure}
        \centering
        \includegraphics[width=0.75\linewidth]{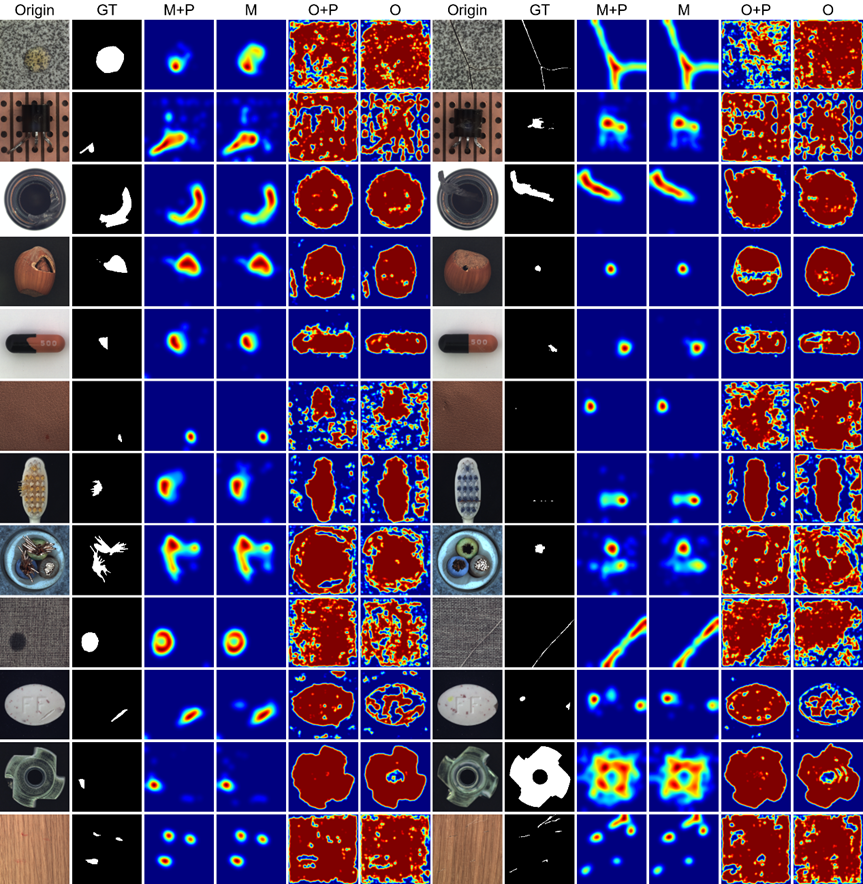}
        \caption{Impact of Decoder configuration and pseudo-labeling head during inference, results on MVTec AD Target Domains (11/1 split): M = memory bank; P = pseudo-labeling; O = original Decoder inference.}
        \label{fig:s7}
    \end{figure}
    \begin{figure}
        \centering
        \includegraphics[width=0.75\linewidth]{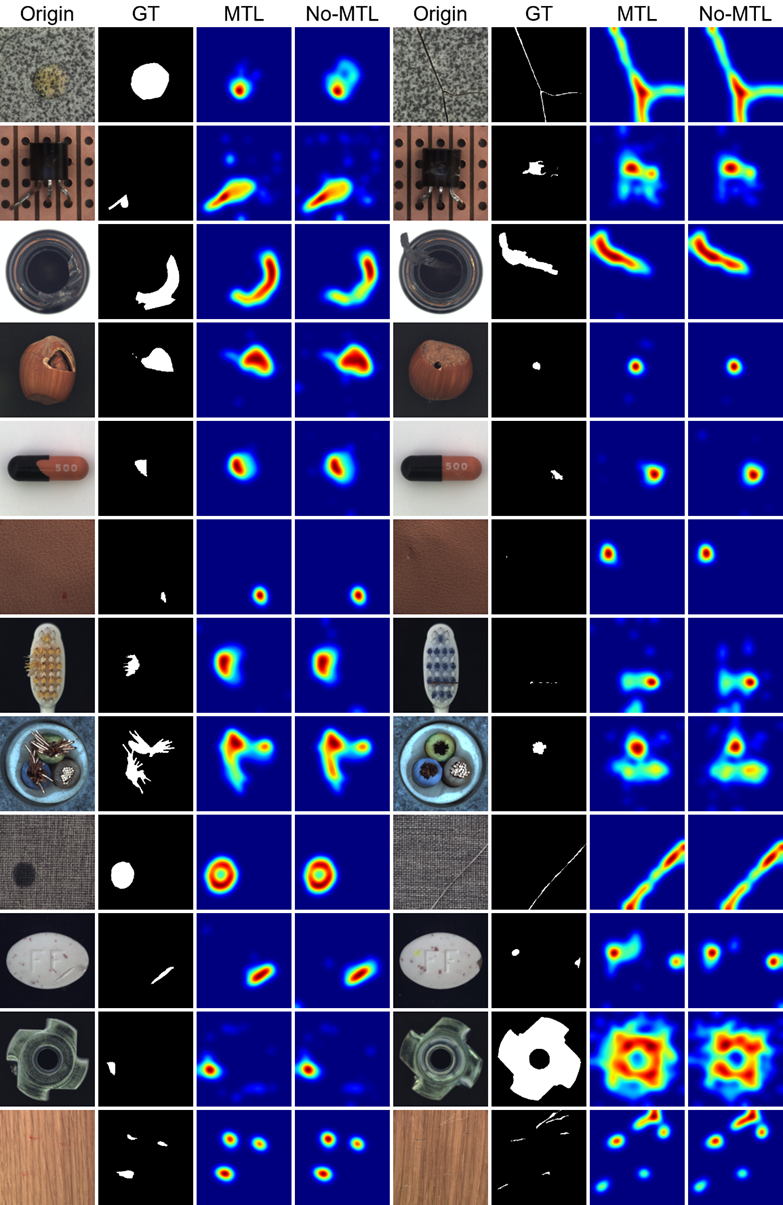}
        \caption{Impact of MTL configuration during training process in the 8/4 split.}
        \label{fig:s8}
    \end{figure}
    \begin{figure}
        \centering
        \includegraphics[width=0.75\linewidth]{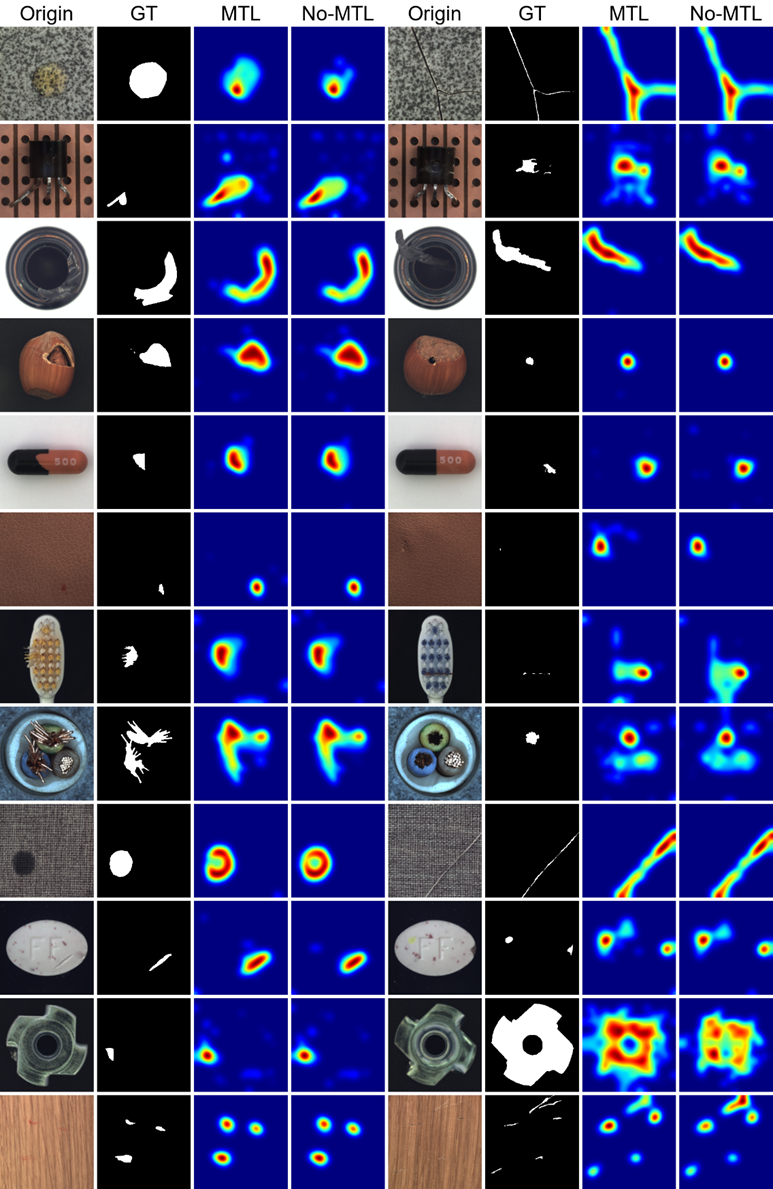}
        \caption{Impact of MTL configuration during training process in the 9/3 Split.}
        \label{fig:s9}
    \end{figure}
    \begin{figure}
        \centering
        \includegraphics[width=0.75\linewidth]{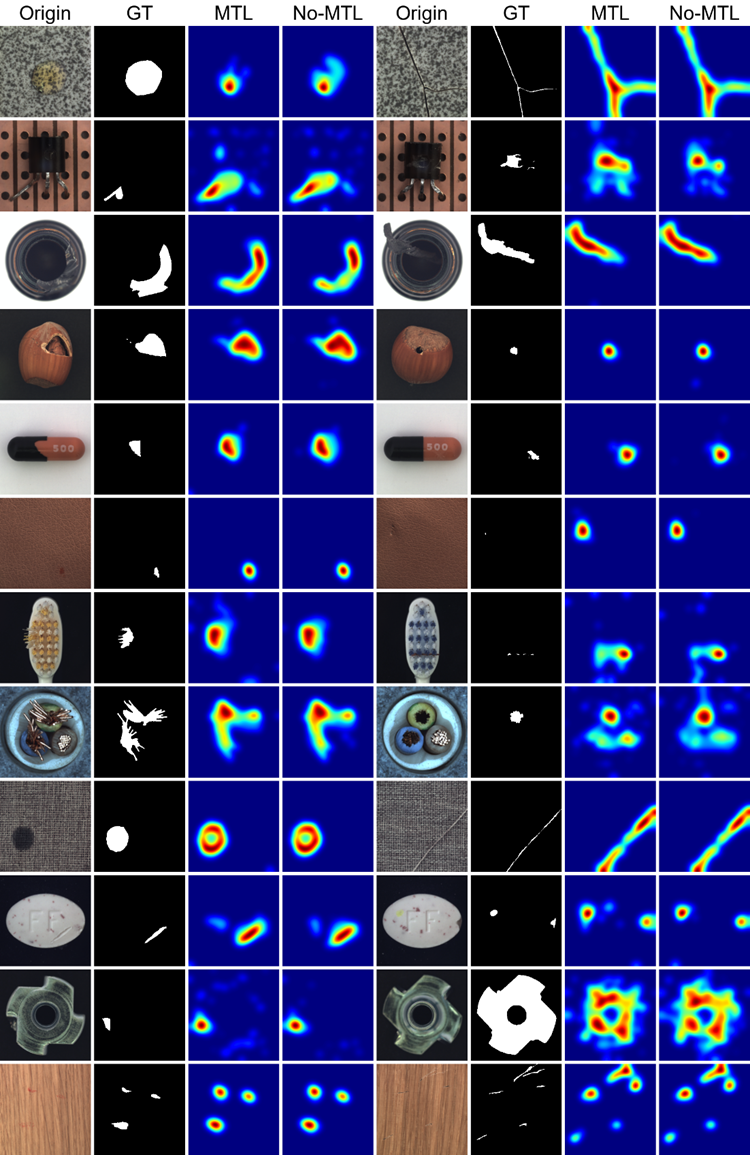}
        \caption{Impact of MTL configuration during training process in the 10/2 split.}
        \label{fig:s10}
    \end{figure}
\begin{figure}
    \centering
    \includegraphics[width=0.75\linewidth]{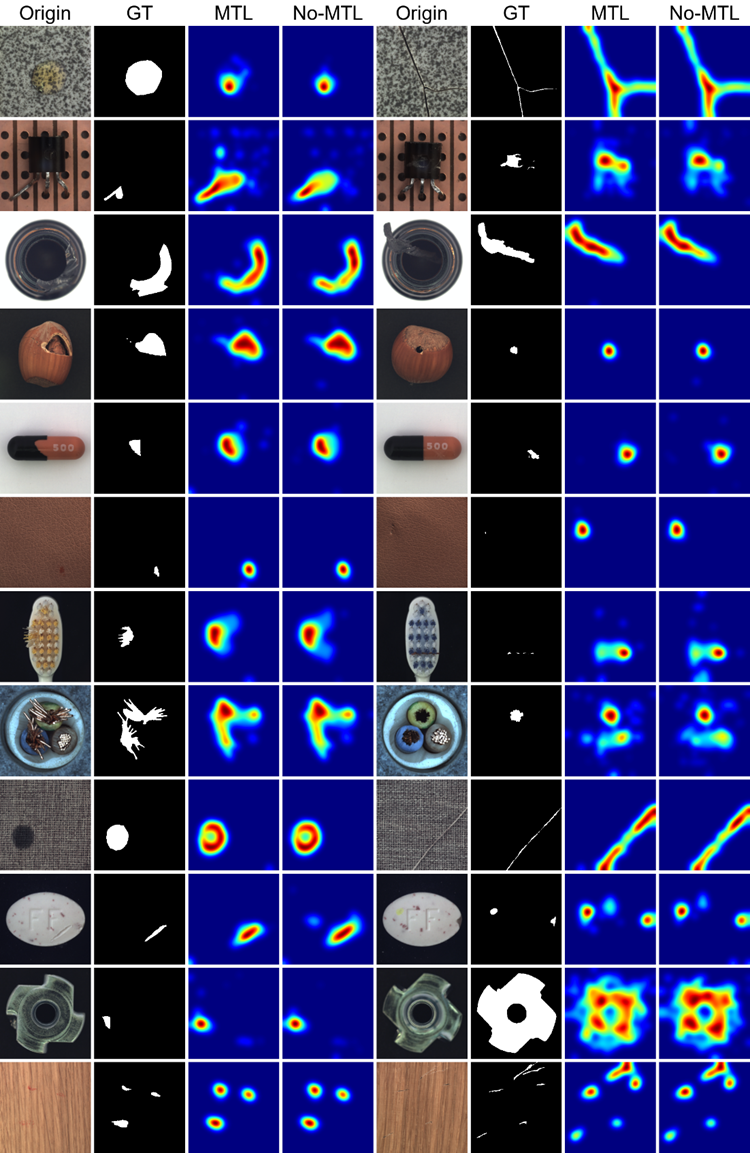}
    \caption{Impact of MTL configuration during training process in the 11/4 split.}
    \label{fig:s11}
\end{figure}
\begin{figure}
    \centering
    \includegraphics[width=1\linewidth]{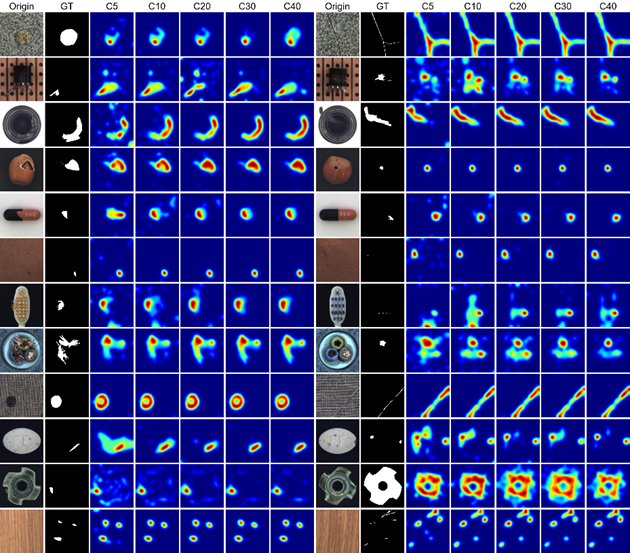}
    \caption{Impact of prototype configuration during inference process in the 8/4 split.}
    \label{fig:s12}
\end{figure}
\begin{figure}
    \centering
    \includegraphics[width=1\linewidth]{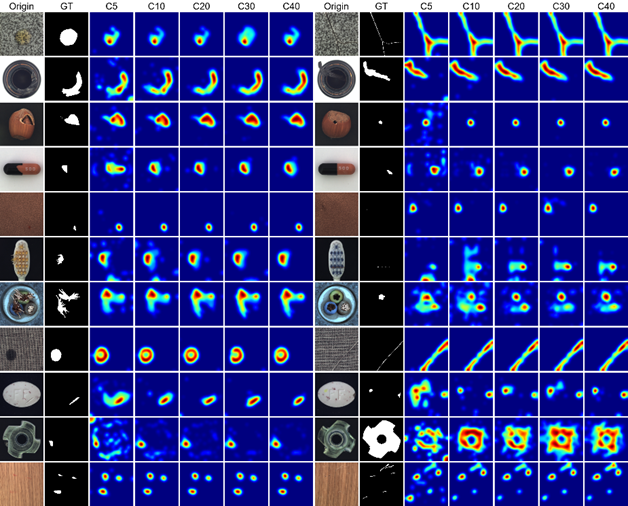}
    \caption{Impact of prototype configuration during inference process in the 9/3 split.}
    \label{fig:s13}
\end{figure}
\begin{figure}
    \centering
    \includegraphics[width=1\linewidth]{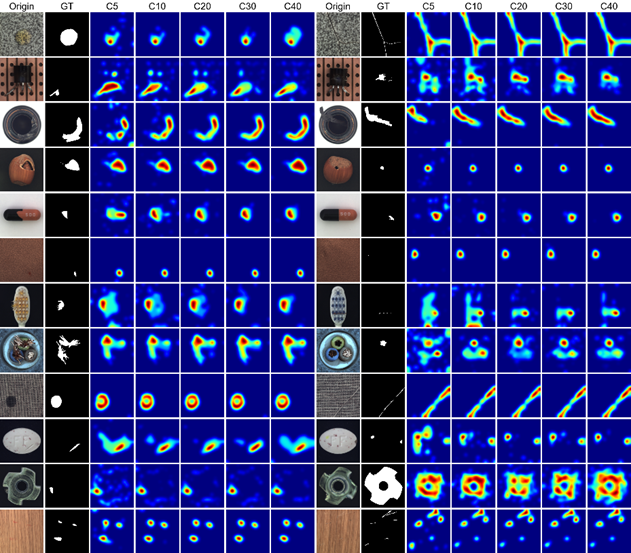}
    \caption{Impact of prototype configuration during inference process in the 10/2 split.}
    \label{fig:s14}
\end{figure}
\begin{figure}
    \centering
    \includegraphics[width=1\linewidth]{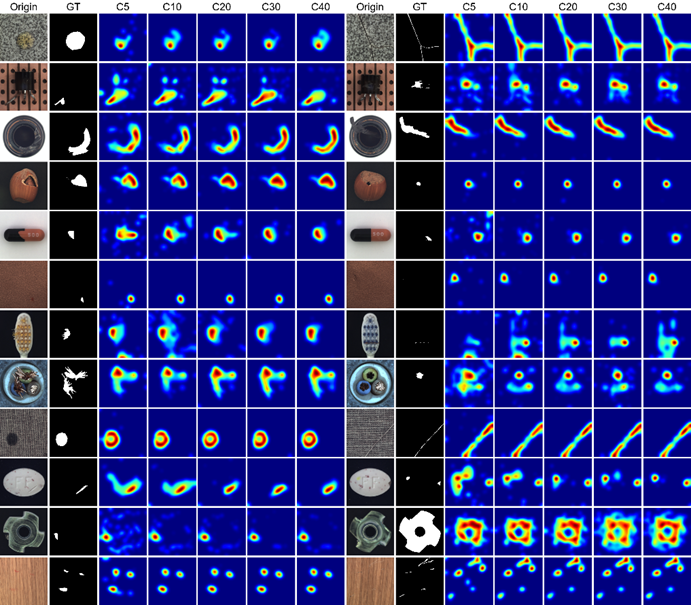}
    \caption{Impact of prototype configuration during inference process in the 11/1 split.}
    \label{fig:s15}
\end{figure}
\end{document}